\documentclass{tlp}
\usepackage{aopmath}
\newcommand{\ignore}[1]{}

\newtheorem{definition}{Definition}[section]
\newtheorem{example}{Example}[section]

\newtheorem{remark}{Remark}[section]

\usepackage{graphicx}
\usepackage{latexsym}
\usepackage[latin1]{inputenc}
\usepackage{color}
\usepackage{multicol}
\usepackage{amssymb,xspace}
\usepackage{subfigure}
\usepackage{enumerate}
\usepackage{hhline}
\usepackage{hyperref}

\usepackage{multicol}
\usepackage{amssymb}
\usepackage{hhline}
\usepackage{fancyvrb}
\usepackage{float}
\usepackage{floatflt}

\newcommand{\boxtheorem}{\hfill $\Box$\\}
\newcommand{\nit}[1]{{\it #1}}
\newcommand{\ul}[1]{\underline{#1}}

\newcommand{\mc}[1]{\mathcal{ #1}}

\newcommand{\mbf}[1]{\mathbf{ #1}}

\newcommand{\mf}[1]{\mathfrak{ #1}}

\newcommand{\msf}[1]{\mathsf{ #1}}

\newcommand{\e}{\mathbf{e}}
\newcommand{\st}{\mbox{\bf \sf s}}
\newcommand{\doo}{\mbox{\bf \sf do}}
\newcommand{\bstar}{\mathbf{\star}}
\newcommand{\ori}{\mbox{\bf \sf o}}

\newcommand{\n}{~{\it not}~}

\newcommand{\iot}{\mathbf{\hat{\iota}}}
\newcommand{\epsi}{\mathbf{\hat{\epsilon}}}

\newcommand{\red}[1]{#1}
\newcommand{\newred}[1]{#1}

\begin{document}

\title[Declarative  Approaches to Counterfactual Explanations]{Declarative  Approaches to Counterfactual Explanations for Classification\thanks{In memory of Prof. Jack Minker (1927-2021), a scientist, a scholar, a visionary; a generous, wise and committed man.}}

\author[L. Bertossi]
{LEOPOLDO BERTOSSI\\Universidad Adolfo Ib\'a\~nez, Faculty of Engineering and Sciences\\
and\\ Millenium Institute on Foundations of Data (IMFD)\\
Santiago, Chile.\\leopoldo.bertossi@uai.cl}

\pagerange{\pageref{firstpage}--\pageref{lastpage}}
\volume{\textbf{10} (3):}
\jdate{March 2002}
\setcounter{page}{1}
\pubyear{2002}

\maketitle

\label{firstpage}

\begin{abstract}
We propose answer-set programs that specify and compute counterfactual interventions on entities that are input on a classification model. In relation to the outcome of the model, the resulting counterfactual entities serve as a basis for the definition and computation of causality-based explanation scores for the feature values in the entity under classification, namely {\em responsibility scores}. The approach and the programs can be applied with black-box models, and also with models that can be specified as logic programs, such as rule-based classifiers. The main focus of this work is on the specification and computation of {\em best}  counterfactual  entities, i.e. those that lead to maximum responsibility scores. From them one can read off the explanations as maximum responsibility feature values in the original entity.  We also extend the programs to bring into the picture semantic or domain knowledge. We show how the approach could be extended by means of probabilistic methods, and how the underlying probability distributions could be modified through the use of constraints. \newred{Several examples of programs written in the syntax of the {\em DLV} ASP-solver,  and run with it, are shown.}
\end{abstract}

\section{Introduction}\label{sec:intro}

\newred{\paragraph{\bf \em Counterfactual Explanations.}
Providing} explanations to results obtained from machine-learning (ML)
models has been recognized as critical in many applications. It has
also become an active research direction in explainable ML \cite{molnar}, and the broader area of {\em explainable AI}. Explanations
become particularly relevant when decisions are automatically made by those
models, possibly with serious consequences for stake holders. Since
most of those models are algorithms
learned from training data, providing explanations may not
be easy or possible. These models are or can be seen as  {\em black-box
models}. Even if the components of the models, say their structure, mathematical functions, and parameters, are relatively clear and accessible, characterizing and measuring the relevance of a particular feature value for the classification of an entity may not be that clear from a sheer inspection of the model.

\ignore{Furthermore, even when the details of the model are
accessible, there is nothing like a universally accepted definition of
what is an explanation for an outcome of the algorithm.}

In AI, explanations have been investigated, among other areas, under {\em actual causality} \cite{Halpern05}, where {\em counterfactual interventions}
on a causal {\em structural model} are central. They are hypothetical updates on the model's
variables, to explore if and how the outcome of the
model changes or not. In this way, explanations for an original output are defined and computed. Counterfactual interventions have been  used, explicitly or implicitly, with ML models, in particular with classification models  \cite{martens,wachter,russell,karimiA,datta,deem}. Counterfactual explanations for query answers from databases have also been investigated \cite{suciu,tocs,foiks18}, an area that is not unrelated to classification results (c.f. Section \ref{sec:compl}).

\red{In this work we start by introducing and formalizing  the notion of {\em counterfactual explanation} for the classification of an entity $\e$ (think of a finite record of feature values) with a certain label $\ell$. Counterfactual explanations are alternative versions, $\e'$, of $\e$ that differ from $\e$ by a set of feature values, but are classified with a different label $\ell'$. In this work, ``best counterfactual explanations" come in two forms, as s-explanations and c-explanations. The former minimize (under set inclusion) the {\em set} of features whose values are changed; while the latter, minimize the {\em number} of changed feature values.}

\newred{\paragraph{\bf \em Explanation Scores and Responsibility.} We use counterfactual explanations} \red{to define the \ $\mbox{\sf x-Resp}$ score, \ an explanation  {\em responsibility score} for a feature value of the entity under classification.} The idea is to identify and compute the {\em most responsible} feature values for the outcome. This score is adapted from  the general notion of responsibility  used in actual causality \cite{Chockler04}.

\red{More specifically, in this work, we concentrate our interest mostly and mainly on specifying and computing, for a given and fixed classified entity $\e$,  {\em all the best explanations} as represented as counterfactual entities (that have feature values changed and a different label), where ``best" for a counterfactual entity means that the number of changes of feature values (with respect to $\e$) takes  a minimum value (the c-explanations mentioned above). At the same time, we are also interested in the {\em maximum-responsibility feature values}. These two goals are directly related: The maximum-responsibility feature values (in the original entity $\e$) are exactly those that appear as changed in a best counterfactual entity.}

\red{The $\mbox{\sf x-Resp}$ score belongs to the family of {\em feature attribution} scores, among which one finds the popular $\mbox{\sf Shap}$ score \cite{lund20}. While $\mbox{\sf Shap}$ is based on the Shapley value that is widely used in game theory, the $\mbox{\sf x-Resp}$ score is based on actual causality.  Experimental results with an extended version (that we briefly describe) of the responsibility score investigated in this work, and comparisons with other scores, in particular, with $\mbox{\sf Shap}$,  are reported in \cite{deem}. However,  only a fully procedural approach to the $\mbox{\sf x-Resp}$ score was followed in \cite{deem}. Furthermore, for performance and comparison-related reasons, the number of counterfactually changeable feature values was bounded a priori. }

\newred{\paragraph{\bf \em Specifying Counterfactuals and Reasoning.}  Answer-set programming} (ASP) is an elegant and powerful logic programming paradigm that allows for declarative specifications of a domain or a problem. From those specifications one can do logical reasoning, and quite usefully, non-monotonic reasoning \cite{asp,GK14}. ASP has been applied to the specification and solution of hard combinatorial problems, and also to different tasks related to databases. Among the latter, we find the specification of repairs of databases w.r.t. integrity constraints, virtual data integration, specification of privacy views \cite{bertossi11,gems}, and causality in databases \cite{foiks18}.

In this work we also introduce {\em Counterfactual Intervention Programs} (CIPs), which are answer-set programs (ASPs)  that specify counterfactual interventions and counterfactual explanations, and also allow to specify and compute the responsibility scores.  \red{More specifically, CIPs are designed to be used to compute: (a) counterfactual explanations (best or not), (b) best counterfactual explanations (i.e. c-explanations), and (c) highest-responsibility features values (in the original entity), that is, with maximum $\mbox{\sf x-Resp}$ score.}

\red{ As already mentioned above, c-explanations lead to maximum responsibility scores, and maximum-responsibility scores are associated to c-explanations. However, in order to compute a maximum-responsibility score (or a feature value which attains  it), we may not need all the best counterfactual entities that ``contain" that particular feature value. Actually, only one of them would be good enough. Still, our CIPs compute all the counterfactual entities, and all (and only) the best counterfactual entities if so required.}

 \red{ In this regard, it is worth mentioning that  {\em counterfactual explanations}, best or not,   are interesting {\em per se}, in that they could show what could be done differently in order to be obtain a different classification \cite{max}. For example, someone, represented as an entity $\e$, who applies for a loan at a bank, and is deemed by the bank's classifier as unworthy of the loan, could benefit from knowing that a reduction of the number of credit cards he/she owns might lead to a positive assessment. This counterfactual explanation would be considered as {\em actionable} (or {\em feasible} or a {\em recourse}) \cite{ustun,karimiB}.}

\red{ Furthermore, having all (or only the best) counterfactual explanations, we could think of doing different kinds of meta-analysis and meta-analytics on top of them. In some cases, query answering on the CIP can be leveraged, e.g. to know if a highest responsibility feature value appears as changed in every best counterfactual entity (c.f. Section \ref{sec:disc}).}

 \red{CIPs can be applied to black-box models that can be invoked from the program; and also with any classifier that can be specified within the program, such as a rule-based classifier. Decision trees for classification can be expressed by means of rules. As recent research shows, other established  classifiers, such as  random forests,  Bayesian networks, \ and neural networks, \ can be compiled into Boolean circuits \cite{shi20,shih18,Choi20,marques19}, opening the possibility of specifying them by means  of ASPs.}

\newred{\paragraph{\bf \em Answer-Set Programming and Semantic Knowledge.} Our declarative }approach to counterfactual interventions is particularly appropriate for bringing into the game additional, declarative, semantic knowledge. \red{As we show, we could easily adopt {\em constraints} when computing counterfactual entities and scores. In particular, these constraints can be used to concentrate only on {\em actionable} counterfactuals (as defined by the constraint). In the loan example, we could omit counterfactual entities or explanations that require from the applicant to reduce his/her age. Imposing constraints in combination with purely procedural approaches is much more complicated. In our case, we can easily and seamlessly  integrate logic-based semantic specifications with declarative programs, and use the generic and optimized solvers behind ASP implementations. In this work we include several examples of CIPs specified in the language of the {\em DLV} system \cite{leone} and its extensions, and run on them. }

\red{We establish that ASPs are the right declarative and computational tool for our problem since they can be used to: (a) Compute all the counterfactual explanations, in particular, the best ones. (b) Provide through each model of the program all the information about a particular counterfactual explanation. (c) Compute maximum-responsibility scores and the counterfactual explanations associated to them. (d) In many cases, specify as a single program the classifier together with the explanation machinery. (d) Alternatively, call a classifier from the program as an external predicate defined in, say Python. (e) Bring into the picture additional, logic-based knowledge, such as semantic and actionability constraints. (f) Do query answering, under the skeptical and brave semantics, to analyze at a higher level the outcomes of the explanation process. Furthermore, and quite importantly, the ASPs we use match the intrinsic data complexity of the problem of computing maximum-responsibility feature values, as we will establish in this work.}

We discuss why and how the responsibility score introduced and investigated early in this work has to be extended in some situations, leading to a generalized probabilistic responsibility score introduced in \cite{deem}. After doing this, we show how logical constraints that capture domain knowledge can be used to modify  the underlying probability distribution on which a score is defined.

\newred{\paragraph{\bf \em Article Structure and Contents.}
This paper is structured} as follows. \red{In Section \ref{sec:backgr} we provide background material on classification models and answer-set programs.} In Section \ref{sec:causes} we introduce counterfactual interventions and the responsibility score. In Section \ref{sec:compl}, we investigate the data complexity of the computation of maximum-responsibility feature values.  In Section \ref{sec:CIPs} we introduce and analyze the CIPs.
 In Section   \ref{sec:sem}, we consider some extensions of CIPs that can be used to capture additional domain knowledge. \red{In Section \ref{sec:new}, and for completeness, we briefly motivate and describe the probabilistic responsibility score  previously introduced in \cite{deem}. In this regard, we also show how to bring logical constraints into the probabilistic setting of the responsibility score.} \red{In Section \ref{sec:relw}, the discuss related work. } In Section \ref{sec:disc}, we conclude with a discussion and concluding remarks.

 \red{\newred{This paper is an  extended version of the conference paper \cite{rr20}.} This current version, in addition to improving and making more precise the presentation in comparison with the conference version, includes a considerable amount of  new material. Section \ref{sec:backgr},  containing background material, is new. In particular, it contains a new example of a decision-tree classifier that is used as an additional running example throughout the paper. Section \ref{sec:compl} on the complexity of $\mbox{\sf x-Resp}$ computation is completely new, so as Section \ref{sec:hcf} that retakes complexity at the light of CIPs. This version also contains several examples of the use of the {\em DLV} system for specifying and running CIPs.  There were no such examples in the conference version. Section \ref{sec:new} is new. It contains two subsections; the first  dealing with non-binary features and the need to extend in probabilistic terms the definition of the $\mbox{\sf x-Resp}$ score; and the second, with the combination of probabilistic uncertainty and domain knowledge. We now have the new Section \ref{sec:relw} on related work. The final Section \ref{sec:disc} has been considerably revised and extended.}

\section{\red{Background}}\label{sec:backgr}

\subsection{\red{Classification models and counterfactuals}}

In a particular domain we may have a finite set $\mc{F}= \{F_1, \ldots, F_n\}$ of {\em features} (a.k.a. as {\em variables}). More precisely, the features $F_i$ are functions that take values in their {\em domains} $\nit{Dom}(F_i)$.\footnote{This is customary parlance, but, since a feature is a function,   in Mathematics we would call this {\em the range} of $F_i$.} As is commonly the case, we will assume that these domains are {\em categorical}, i.e.  finite and without any a priori order structure. These features are used to describe or represent {\em entities} in an {\em underlying population} (or application domain),
\ignore{$\mathbf{E}$,} as records (or tuples) $\e$ formed by the values \ignore{, $F_i(\mbf{E})$,} the features take on the entity\ignore{ $\mathbf{E}$}. Actually, with a bit of abuse of notation, we represent entities directly as these records of feature values: \ $\e = \langle F_1(\mathbf{e}), \ldots, F_n(\e)\rangle$.  A feature $F$ is said to be {\em Boolean} or {\em propositional} if $\nit{Dom}(F) = \{\msf{true}, \msf{false}\}$, for which we sometimes use $\{\msf{yes}, \msf{no}\}$ or simply, $\{1,0\}$.

For example, if the underlying population contains people, and then, each entity represents a person, features could be $\mc{F} = \{\msf{Weight},$ $\msf{Age}, \msf{Marrid}, \msf{EdLevel}\}$, with domains: $\nit{Dom}(\msf{Weight}) =
\{\msf{overweight}, \msf{chubby}, \msf{fit},$ $\msf{slim},$ $ \msf{underweight}\}$; $\nit{Dom}(\msf{Age}) =
\{\msf{old}, \msf{middle}, \msf{young},$ $\msf{child}\}$; $\nit{Dom}(\msf{Married}) =
\{\msf{true}, \msf{false}\}$; $\nit{Dom}(\msf{EdLevel}) =
\{\msf{low},$ $ \msf{medium}, \msf{high},$ $\msf{top}\}$. \ A particular  entity could be (represented by) $\e = \langle \msf{fit}, \msf{young},$ $ \msf{false}, \msf{top}\rangle$. Here, $\msf{Married}$ is a propositional (or binary) feature.

A {\em classifier}, $\mc{C}$, for a domain of entities $\mc{E}$ is a computable function $\mc{C}: \mc{E} \rightarrow L$, where $L = \{\ell_1,\ldots, \ell_k\}$ is a set of {\em labels}. We classify the entity by assigning a label to it. We do not have to assume any order structure on $L$. In the example above, we may want to classify entities in the people's population according to how appropriate they are for military service. The set of labels could be $L = \{\msf{good}, \msf{maybe}, \msf{noway}\}$, then the classifier could be such, that $\mc{C}(\langle \msf{fit}, \msf{young},$ $ \msf{false}, \msf{top}\rangle) = \msf{good}$, but $\mc{C}(\langle \msf{overweight}, \msf{old},$ $ \msf{true}, \msf{low}\rangle) = \msf{noway}$. \ When the set of labels has two values, typically, $\{\msf{yes}, \msf{no}\}$, we have a {\em binary or Boolean classifier}. In this work we will consider mostly binary classifiers.

\begin{example} \label{ex:tree} This is a popular example taken from \cite{mitchell}. Consider the set of features $\mc{F} = \{\mathsf{Outlook}, \mathsf{Humidity}, \mathsf{Wind}\}$, with $\nit{Dom}(\mathsf{Outlook}) =$ $\{\mathsf{sunny}, \mathsf{overcast},
\mathsf{rain}\}$,  $\nit{Dom}(\mathsf{Humidity}) =$ $\{\mathsf{high}, \mathsf{normal}\}$, $\nit{Dom}(\mathsf{Wind}) =$ $\{\mathsf{strong},$ $ \mathsf{weak}\}$. An entity under classification has a value for each of the features, e.g. $\e = \nit{ent}(\mathsf{sunny}, \mathsf{normal}, \mathsf{weak})$. The problem is about deciding about playing tennis or not under the conditions represented by that entity, which can be captured as a classification problem, with labels $\mathsf{yes}$ or $\mathsf{no}$ (sometimes we will use $1$ and $0$, resp.).\vspace{5mm}

\begin{center}\begin{tabular*}{7cm}{|c|}\cline{1-1}\\
\phantom{o} \hspace{6cm} \phantom{o}\\
\phantom{o}\\
\phantom{o}\\
\phantom{o}\\
\phantom{o}\\
\phantom{o}\\
\phantom{o}\\
\phantom{o}\\
\cline{1-1}
\end{tabular*}
\end{center}

\vspace{-4.6cm}
\begin{figure}[h]
\begin{center}
\includegraphics[width=6cm]{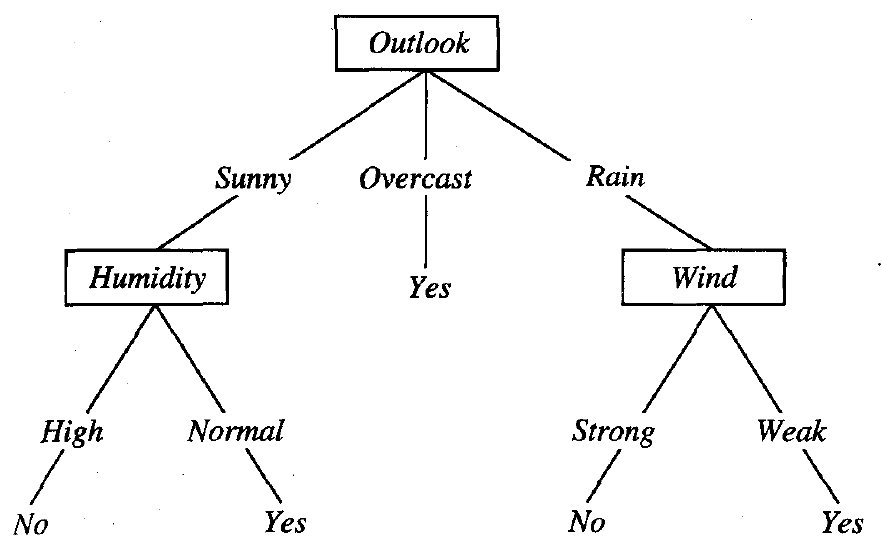}
\caption{A Decision Tree}\label{fig:tree}
\end{center}
\end{figure}
The Boolean classifier is given as a decision-tree, as shown in Figure \ref{fig:tree}. The decision is computed by following the feature values along the branches of the tree.  The entity $\e$ at hand gets label $\msf{yes}$.
\boxtheorem
\end{example}


Of course, a decision tree could be just arbitrarily given, but the common situation is {\em to learn} it from training examples, that is from a collection of entities that come already with an assigned label. In more general terms, building a classifier, $\mc{C}$, from a set of training data, i.e. a set  of pairs $T = \{\langle \e_1,\ell(\e_1)\rangle,$ $ \ldots, \langle \e_M,\ell(\e_M)\rangle\}$, with $\ell(\e_i) \in L$,  is about learning a (computational) model for the label function $L$ for the entire domain of entities, beyond $T$. We say that $L$ ``represents" the classifier $\mc{C}$. This {\em supervised learning} task is one of the most common in machine learning. (C.f. \cite{mitchell} or \cite{flach} for great introductions to the subject.)
\ Classifiers may take many different internal forms. They could be decision trees, random forests, rule-based classifiers, logistic regression models, neural network-based classifiers, etc.

In this work we are {\em not} concerned with learning classifiers. We will assume the classifier is given as an input/output computable relation, or that we know how to specify it, for example, trough a logic program. By the same token, we are {\em not} dealing here with any kind of program learning.

Some classes of classifiers  are more ``opaque" than others, i.e. with a more complex and less interpretable internal structure and results \cite{molnar}. Hence, the need for explaining the classification outcomes. The decision tree above is clearly a computable function. Since we have the classifier at hand with an explicit and clear specification, we would consider this classifier to be {\em interpretable}. In this direction, if we obtain a label for a particular entity at hand, we can inspect the model and explain {\em why} we obtained that label, and we could identify the feature values that were relevant for the outcome.

Instead of a decision tree as above,  we could have, for example,  a very complex neural network. Such a model would be much more difficult to interpret, or use to explain why we got a particular output for a particular input. This kind of models are considered to be {\em black-box} models (or al least, {\em opaque}) \cite{rudin}.

Assume for a moment that we do not have the explicit classifier in Example \ref{ex:tree}, but we interact only with the box that contains it. We input entity  $\e = \nit{ent}(\mathsf{sunny}, \mathsf{normal}, \mathsf{weak})$, and we obtain the output $\msf{yes}$. We want to know what are the feature values in $\e$ that influenced the outcome the most. Actually, we want to get a {\em numerical score} for each of the entity's feature values, in such a way that the higher the score, the more relevant is the feature value for the outcome.

Different scores may be defined. A popular one is {\sf Shap} \cite{lund20}, which has been investigated in detail for some classes of classifiers \cite{aaai21,guy}. In this work we concentrate on $\mbox{\sf x-Resp}$,  a {\em responsibility score} that was introduced, in a  somewhat {\em ad hoc} manner,   in \cite{deem}. In this work, we present $\mbox{\sf x-Resp}$ in a causal setting, on the basis of counterfactual interventions and their responsibilities, following the general approach to causal responsibility in \cite{Chockler04}.

Just for the gist, and at the light of the example at hand, we want to detect and quantify the relevance (technically, the responsibility) of a feature value in $\e = \nit{ent}(\mathsf{sunny}, \underline{\mathsf{normal}}, \mathsf{weak})$, say for feature $\msf{Humidity}$ (underlined), by {\em hypothetically intervening} its value; in this case, changing it from $\msf{normal}$ to $\msf{high}$, obtaining a new entity $\e' = \nit{ent}(\mathsf{sunny}, \underline{\mathsf{high}}, \mathsf{weak})$, a {\em counterfactual version} of $\e$. If we input this entity into the classifier, we now obtain the label $\msf{no}$. This is an indication that the original feature value for $\msf{Humidity}$ is indeed relevant for the original classification. Through numerical aggregations over the outcomes associated to the alternative, counterfactually intervened entities, we can define and compute the $\mbox{\sf x-Resp}$ score. The details can be found in Section \ref{sec:causes}. As mentioned in Section \ref{sec:intro}, these counterfactual entities are also interesting {\em per se}, and not only as a basis for the definition and computation of {\em feature attribution scores}.

\subsection{\red{Answer-set programming}}

As customary, when we talk about {\em answer-set programs}, we refer to {\em disjunctive Datalog programs with weak negation and stable model semantics} \cite{GL91}.  Accordingly, an answer-set program $\Pi$ consists of a finite number of rules of the form

\vspace{1mm}
\begin{equation}
A_1 \vee \ldots \vee A_n \leftarrow P_1, \ldots, P_m, \n N_1, \ldots, \n N_k, \label{eq:rule}
\end{equation}

\vspace{1mm}\noindent
 where $0\leq n,m,k$, and $A_i, P_j, N_s$ are (positive) atoms, i.e. of the form $Q(\bar{t})$, where $Q$ is a predicate of a fixed arity, say, $\ell$, and $\bar{t}$ is a sequence of length $\ell$ of variables or constants.
In rule (\ref{eq:rule}), $A_1, \ldots, \n N_k$ are called {\em literals}, with $A_1$ {\em positive}, and $\n  N_k$, {\em negative}. All the variables in the $A_i, N_s$ appear among those
in the $P_j$.   The left-hand side of a rule is called the {\em head}, and the right-hand side, the {\em body}.  A rule can be seen as a (partial) definition of the predicates in the head (there may be other rules with the same predicates in the head).

The constants in  program $\Pi$ form the (finite) Herbrand universe $H$ of the program. The ground version of
program $\Pi$, $\nit{gr}(\Pi)$, is obtained by instantiating the variables in $\Pi$ in all
possible ways  using
values from $H$. The Herbrand base, $\nit{H\!B}$, of $\Pi$ contains all the atoms obtained as instantiations of
predicates in $\Pi$ with constants in $H$.

A subset $M$ of $\nit{HB}$ is a model of $\Pi$ if it satisfies $\nit{gr}(\Pi)$, i.e.: For every
ground rule $A_1 \vee \ldots \vee A_n$ $\leftarrow$ $P_1, \ldots, P_m,$ $\n N_1, \ldots,
\n N_k$ of $\nit{gr}(\Pi)$, if $\{P_1, \ldots, P_m\}$ $\subseteq$ $M$ and $\{N_1, \ldots, N_k\} \cap M = \emptyset$, then
$\{A_1, \ldots, A_n\} \cap M \neq \emptyset$. $M$ is a minimal model of $\Pi$ if it is a model of $\Pi$, and $\Pi$ has no model
that is properly contained in $M$. $\nit{MM}(\Pi)$ denotes the class of minimal models of $\Pi$.
Now, for $S \subseteq \nit{HB}(\Pi)$, transform $\nit{gr}(\Pi)$ into a new, positive program $\nit{gr}(\Pi)^{\!S}$ (i.e.\  without $\nit{not}$), as follows:
Delete every rule  $A_1 \vee \ldots \vee A_n \leftarrow P_1, \ldots,P_m, \n N_1,$ $ \ldots,
\n N_k$ for which $\{N_1, \ldots, N_k\} \cap S \neq \emptyset$. Next, transform each remaining rule $A_1 \vee \ldots \vee A_n \leftarrow P_1, \ldots, P_m,$ $\n N_1, \ldots,
\n N_k$ into $A_1 \vee \ldots \vee A_n \leftarrow P_1, \ldots, P_m$. Now, $S$ is a {\em stable model} of $\Pi$ if $S \in \nit{MM}(\nit{gr}(\Pi)^{\!S})$.
Every stable model of $\Pi$ is also a minimal model of $\Pi$. Stable models are also commonly called  {\em answer sets}, and so are we going to do most of the time.

 A program is {\em unstratified} if there is a cyclic, recursive definition of a predicate that involves negation. For example, the program consisting of the rules $a \vee b \leftarrow c, \n d$; \ $d \leftarrow e$, and $e \leftarrow b$ is unstratified, because there is a negation in the mutually recursive definitions of $b$ and $e$. The program in Example \ref{ex:hcf} below is not unstratified, i.e. it is {\em stratified}. A good property of stratified programs is that the models can be upwardly computed following {\em strata} (layers) starting from the {\em facts}, that is from the ground instantiations of rules with empty bodies (in which case the arrow is usually omitted). We refer the reader to \cite{GK14} for more details.

 Query answering under the ASPs comes in two forms. Under the {\em brave semantics}, a query posed to the program obtains as answers those that hold in {\em some} model of the program. However, under the {\em skeptical} (or {\em cautious}) semantics, only the answers that simultaneously hold in {\em all} the models are returned. Both are useful depending on the application at hand.

We will use disjunctive programs. However, sometimes it is possible to use instead {\em normal programs}, which do not have disjunctions in rule heads, and with the same stable models,  in the sense that disjunctive rules can be transformed into a set of non-disjunctive rules. More precisely, the rule in (\ref{eq:rule}) can be transformed into the rules:
\begin{eqnarray*}
A_1 &\leftarrow& P_1, \ldots, P_m, \n N_1, \ldots, \n N_k, \n A_2, \ldots, \n A_n\\
&\cdots&\\
A_n  &\leftarrow& P_1, \ldots, P_m, \n N_1, \ldots, \n N_k, \n A_1, \ldots, \n A_{n-1}.
\end{eqnarray*}
This {\em shift operation}  is possible if the original program is {\em head-cycle free} \cite{Ben94,dantsin}, as we define now.  The {\em dependency graph} of a program $\Pi$, denoted  $\nit{DG}(\Pi)$, is a
directed graph whose nodes are the atoms of the associated ground program $\nit{gr}(\Pi)$. There is
an arc from $L_1$ to $L_2$
iff there is a rule in $\nit{gr}(\Pi)$ where $L_1$ appears positive in the body and $L_2$ appears in
the head. $\Pi$ is {\em head-cycle free} (HCF) iff $\nit{DG}(\Pi)$ has no cycle through two atoms that belong to the head of a same
rule.

\begin{example} \label{ex:hcf} Consider the following  program $\Pi$ that is already ground.

\begin{multicols}{2}
\begin{eqnarray*}
a \vee b &\leftarrow& c\\
d &\leftarrow& b\\
a \vee b &\leftarrow& e, \ {\it not} {\it f}\\
e &\leftarrow&
\end{eqnarray*}

\phantom{o}

The program has two stable models: \ $S_1 = \{e, a\}$ and $S_2 = \{e,b,d\}$.

Each of them expresses that the atoms in it are true, and any other atom that does not belong to it, is false.
\end{multicols}

These models are incomparable under set inclusion, and they are minimal models in that any proper subset of any of them is not a model of the program.

\begin{figure}[h]
\includegraphics[width=5cm]{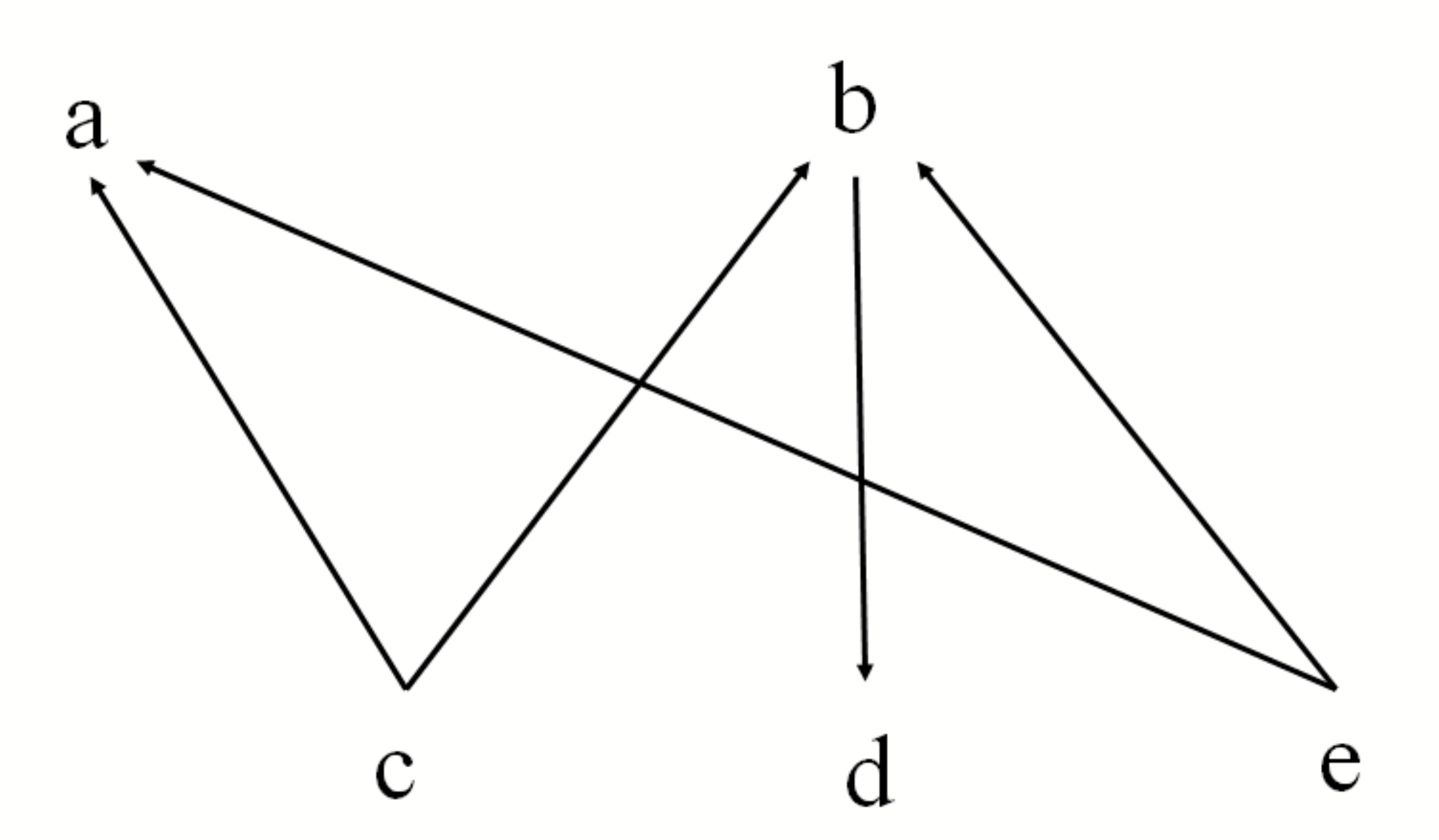}
\caption{Dependency graph $\nit{DG}(\Pi)$} \label{fig:hcf}
\end{figure}
The dependency graph is shown in Figure \ref{fig:hcf}. We can see that the program $\Pi$ is head-cycle free, because there is no cycle involving both $a$ and $b$, the atoms that appear in the disjunctive head.
As a consequence, the program can be transformed into the following non-disjunctive program

\vspace{-7mm}
\begin{multicols}{2}
\begin{eqnarray*}
a &\leftarrow& c, \ \n b\\
b &\leftarrow& c, \ \n a\\
d &\leftarrow& b
\end{eqnarray*}

\begin{eqnarray*}
a &\leftarrow& e, \ {\it not} {\it f}, \ \n b\\
b &\leftarrow& e, \ {\it not} {\it f}, \ \n a.\\
e &\leftarrow&
\end{eqnarray*}
\end{multicols}
This program has the same stable models as the original one.
\boxtheorem
\end{example}

We will return to HCF-programs in Section \ref{sec:hcf}.

\section{Counterfactual Explanations and the {\sf x-Resp} Score}\label{sec:causes}

We consider {\em classification models}, $\mc{C}$, that are  represented by an input/output relation. Inputs are the so-called {\em entities}, $\e$, which  are represented as records, $\e = \langle x_1, \ldots, x_n\rangle$, where $x_i$ is the value $F_i(\e) \in \nit{Dom}_i := \nit{Dom}(F_i)$ taken  on $\e$ by the {\em feature}  $F_i \in \mc{F} = \{F_1, \ldots, F_n\}$. The entities form a {\em population} $\mc{E}$. The output of a classifier is represented by a {\em label function} $L$ that maps entities $\e$ to $0$ or $1$, the binary result of the classification. That is, to  simplify the presentation, we concentrate here on binary classifiers, but this is not essential.  Furthermore, we  consider domains $\nit{Dom}(F_i)$ with a finite number of categorical values.

 In this work, we are not assuming that we have  an explicit classification model, and we do not need it. All we need is to be able to invoke  it. It could be a ``black-box" model.

The problem is the following: Given an entity $\mbf{e}$ that has received the label $L(\mbf{e})$, provide an ``explanation" for this outcome. In order to simplify the presentation, and without loss of generality, {\em we  assume that label $1$ is the one that has to be explained}. It is the ``negative" outcome one has to justify, such as the rejection of a loan application.

\red{Counterfactual explanations are defined in terms of {\em counterfactual interventions} that simultaneously change
feature values in $\e$, in such a way that the updated record gets a new label. A {\em counterfactual explanation} for the classification of
$\e$ is, then, an alternative entity $\e'$ that is classified with a label different from that of $\e$. In general, we are interested in counterfactual explanations that are more informative about the role of the feature values in $\e$, which lead to its obtained label. These are the entities that are obtained from $\e$ through a  {\em minimal counterfactual interventions}. } Minimality can be defined in different ways, and we adopt an abstract approach, assuming  a partial order relation $\preceq$ on counterfactual interventions.

\begin{definition} \label{def:causal:explanation}  Consider a binary classifier represented by its label function $L$, and a fixed input entity $\e = \langle x_1, \ldots, x_n\rangle$, with $F_i(\e) = x_i$, \ $1 \leq i \leq n$, and $L(\e) =1$.

\noindent (a) \ An {\em intervention}  $\iot$ on $\e$ is a set of the form $\{\langle F_{i_1},x_{i_1}'\rangle, \ldots, \langle F_{i_K},x_{i_K}'\rangle\}$, with $F_{i_s} \neq F_{i_\ell}$, for $s\neq \ell$,  \ $x_{i_s} \neq x_{i_s}' \in \nit{Dom}(F_{i_s})$. \ We denote with $\iot(\e)$ the record obtained by applying to $\e$ intervention $\iot$, i.e. by replacing in $\e$ every $x_{i_s} = F_{i_s}(\e)$, with $F_{i_s}$ appearing in $\iot$, by $x_{i_s}'$.

\noindent (b) A {\em counterfactual intervention} on $\e$ is an intervention $\iot$ on $\e$, such that $L(\iot(\e)) = 0$. \ \red{The resulting entity $\e' = \iot(\e)$ is called a {\em counterfactual entity} (for $\e$).}

\noindent \red{(c) A {\em $\preceq$-minimal} counterfactual intervention $\iot$ is such that there is no counterfactual intervention $\iot'$ on $\e$ with
$\iot' \prec \iot$ \ (i.e. \ $\iot' \preceq \iot$, but not  $\iot \preceq \iot'$). \ The resulting entity $\e' = \iot(\e)$ is called a {\em $\preceq$-minimal counterfactual entity}.}

\noindent (d) \red{A {\em counterfactual explanation}} for $L(\e)$ is a set  of the form $\epsi = \{\langle F_{i_1},
x_{i_1}\rangle, \ldots,$  $ \langle F_{i_K}, x_{i_K}\rangle\}$, \red{with $F_{i_j}(\e) = x_{i_j}$}, for which there is a counterfactual intervention \red{$\iot(\epsi) = $ $\{\langle F_{i_1},
x_{i_1}'\rangle, \ldots,$ $ \langle F_{i_K},x_{i_K}'\rangle\}$ for $\e$.} 

\noindent \red{(e) A counterfactual explanation $\epsi$ for $L(\e)$ is {\em $\preceq$-minimal} if its associated counterfactual intervention $\iot(\epsi)$ is  a $\preceq$-minimal counterfactual intervention  on $\e$.} \boxtheorem

\end{definition}

\vspace{-5mm}
\red{\begin{remark}\label{rem:loose}
Counterfactual explanations contain feature values of the original entity $\e$. They contain relevant information about the classification result, and interventions are used to characterize and identify them. For this reason, we will usually call an alternative entity $\e'$ obtained from $\e$ through a counterfactual intervention, a {\em counterfactual explanation} as well: The counterfactual explanation in the sense of  Definition \ref{def:causal:explanation}(d) can be read-off from $\e'$. \boxtheorem
\end{remark}}

 Several minimality criteria can be expressed in terms of partial orders. We will explicitly consider two of them.

 \begin{definition} \label{def:min} \red{ Let  $\iot_1$ and $\iot_2$ be interventions on an entity $\e$. \
 (a) \ $\iot_1 \leq^s \iot_2$ iff $\pi_{1}(\iot_1) \subseteq \pi_{1}(\iot_2)$, with $\pi_{1}(\iot)$ the projection of $\iot$ on the first position. \ (b) \ $\iot_1 \leq^c \iot_2$ iff $|\iot_1| \leq |\iot_2|$.} \boxtheorem
\end{definition}
\ignore{ A maybe more exotic one when feature values are numerical: \ $\iot_1 \leq^{\mbox{\scriptsize sum}} \iot_2$ iff $\Sigma_{\langle i_j;
x_{i_j},x_{i_j}'\rangle \in \iot_1} |x_{i_j} - x_{i_j}'| \leq  \Sigma_{\langle i_j;
x_{i_j},x_{i_j}'\rangle \in \iot_2} |x_{i_j} - x_{i_j}'|$.}
\red{This definition introduces minimality
under set inclusion and cardinality, resp.  The former minimizes -under set inclusion- the set of features whose values are changed. The latter, the number of features that see their values changed. In the following, we will consider only these minimality criteria, mostly the second (c.f. Section \ref{sec:disc} for a discussion).}

\begin{example} \label{ex:first} Consider three binary features $\mc{F} = \{F_1, F_2, F_3\}$, i.e. they take values $0$ or $1$. The  input/output relation of a classifier $\mc{C}$ is shown in Table 1. Let $\e$ be $\e_1$ in the table. We want counterfactual explanations for its label $1$. Any other record in the table can be seen as the result of an intervention on $\e_1$.  However, only $\e_4, \e_7, \e_8$ are (results of) counterfactual interventions in that they switch the label to $0$.

\begin{multicols}{2}
\begin{center}
\begin{tabular*}{4cm}{|c|| c|c|c||c|}\cline{1-5}
entity (id) & $F_1$ & $F_2$ & $F_3$ & $L$\\ \cline{1-5}
$\e_1$ & 0 & 1 & 1 &1\\ \cline{1-5}
$\e_2$ & 1 & 1 & 1 &1\\
$\e_3$ & 1 & 1 & 0 &1\\
$\e_4$ & 1 & 0 & 1 &0\\
$\e_5$ & 1 & 0 & 0 &1\\
$\e_6$ & 0 & 1 & 0 &1\\
$\e_7$ & 0 & 0 & 1 &0\\
$\e_8$ & 0 & 0 & 0 &0\\ \cline{1-5}
\end{tabular*}\\ \vspace{1mm}
{\bf Table 1: \ Classifier \ $\mc{C}$}
\end{center}

For example, $\e_4$ corresponds to the intervention $\iot_4 = \{\langle F_1,1\rangle, \langle F_2,0\rangle\}$, in that $\e_4$ is obtained from $\e_1$ by changing the values of $F_1, F_2$ into $1$ and $0$, resp. For $\iot_4$,  $\pi_{1}(\iot_4) = \{\langle F_1\rangle,\langle F_2\rangle\}$. From $\iot_4$ we obtain the
 counterfactual explanation $\epsi_4 = \{\langle F_1, 0\rangle, \langle F_2, 1\rangle\}$, telling  us that the values $F_1(\e_1) = 0$ and $F_2(\e_1) = 1$ are the joint cause for $\e_1$ to be classified as $1$.

\end{multicols}

\vspace{-3mm} There are three counterfactual explanations: \ $\epsi_4 := \{\langle F_1,0\rangle,\langle F_2,1\rangle\}$, $\epsi_7 :=$ \linebreak $ \{\langle F_2,1\rangle\}$,
and $\epsi_8 := \{\langle F_2,1\rangle, \langle F_3,1\rangle\}$.   \ Here, $\e_4$ and $\e_8$ are incomparable under $\preceq^s$, \ $\e_7 \prec^s \e_4$, \ $\e_7 \prec^s \e_8$, and  $\epsi_7$ turns out to be $\preceq^s$- \ and \ $\preceq^c$-minimal (actually, minimum). \boxtheorem
\end{example}

\vspace{-3mm}Notice that what matters here is {\em what} is intervened, and not {\em how}. By taking a projection, the partial order $\preceq^s$ does not care about the values that replace the original feature values, as long as the latter are changed. \ Furthermore, given $\e$, it would be good enough to indicate the features whose values are relevant, e.g. $\epsi_7 = \{F_2\}$ in the previous example. However, the introduced notation emphasizes the fact that the original values are those we concentrate on when providing explanations.

Every $\preceq^c$-minimal explanation is also $\preceq^s$-minimal. However, it is easy to produce an example showing that a $\preceq^s$-minimal explanation may not be $\preceq^c$-minimal. \ignore{
Until further notice we will concentrate only on $\preceq^s$- and $\preceq^c$-minimal explanations; actually more on the latter.}

\ignore{\begin{definition} \em Given a binary classifier represented by its label function $L$, and a fixed input record $\e$: \
(a) An {\em s-explanation} for $L(\e)$ is a $\preceq^s$-minimal causal explanation for $L(\e)$. \
(b) A {\em c-explanations} $L(\e)$ is a $\preceq^s$-minimal causal explanation for $L(\e)$. \boxtheorem
\end{definition} }

\vspace{1mm}\noindent {\bf Notation:} \ An {\em s-explanation} for $L(\e)$ is a $\preceq^s$-minimal \red{counterfactual explanation} for $L(\e)$. \
A {\em c-explanation} for $L(\e)$ is a $\preceq^c$-minimal \red{counterfactual explanation} for $L(\e)$. \red{So as prescribed in Remark \ref{rem:loose}, we will usually use the terms s-explanation and c-explanation to refer to
the  alternative, intervened entities $\e'$ that are associated to
s- and c-explanations in the sense of Definition \ref{def:min}.} \boxtheorem

\vspace{1mm}
 \red{So far, we have characterized  explanations as sets of (interventions on) features. However, one would also like to quantify the ``causal strength" of a {\em single feature value} in a record representing an entity. Something similar has been done for a single tuple in a database as a cause for a query answer \cite{suciu}, or for a single attribute value in a database tuple \cite{tocs,foiks18}. \ Different {\em scores} for feature values have been proposed in this direction, e.g. $\mbox{\sf Shap}$ in \cite{lund20} and {\sf Resp} in \cite{deem}.  Following \cite{Chockler04}, we will now formulate the latter  as the {\em responsibility} of a feature value as an {\em actual cause} \cite{Halpern05} for the observed classification.}

\begin{definition} \label{def:er}  Consider   an entity $\e$ represented as a record of feature values $x_i=F_i(\e)$, $F_i \in \mc{F}$.

\noindent  (a) \ A feature value $v =F(\e)$, with $F\in \mc{F}$, is a {\em value-explanation}  for $L(\e)$ if there is an s-explanation $\epsi$ for $L(\e)$, such that $\langle F,v\rangle \in \epsi$.

\noindent (b) The {\em explanatory responsibility} of a value-explanation $v =F(\e)$ is:
$$\mbox{\sf x-Resp}_{\e,F}(v):= \nit{max}\{\frac{1}{|\epsi|} : \epsi \ \mbox{ is  s-explanation with } \langle F,v \rangle \in \epsi\}.$$
(c) If $v =F(\e)$ is not a value-explanation, $\mbox{\sf x-Resp}_{\e,F}(v):=0$.
\boxtheorem
\end{definition}

\vspace{-4mm}Notice that (b) can be stated as \ $\mbox{\sf x-Resp}_{\e,F}(v):= \frac{1}{|\epsi^\star|}$, with \
$ \epsi^\star = \mbox{\sf argmin} \{|\epsi|~:~\epsi$ $\mbox{is s-explanation with } \langle F,v\rangle \in \epsi\}$.

 Adopting the terminology of actual causality \cite{Halpern05}, a {\em counterfactual value-explanation} for $\e$'s classification is a value-explanation $v$ with  $\mbox{\sf x-Resp}_{\e,F}(v)=1$. That is, it suffices, without company from other feature values in $\e$, to explain the classification. \ Similarly, an {\em actual value-explanation} for $\e$'s classification is a value-explanation $v$ with
$\mbox{\sf x-Resp}_{\e,F}(v) > 0$. That is, $v$ appears in an s-explanation $\epsi$, say as $\langle F, v\rangle$, but possibly in company of other feature values. In this case, $\epsi \smallsetminus \{\langle F,v \rangle \}$ is  a {\em contingency set} for $v$ \ (c.f. \cite{Halpern05}, and  \cite{suciu} for the case of databases).

\begin{example} (example \ref{ex:first} continued) \  $\epsi_7$, equivalently entity $\e_7$, is the only c-explanation for  $\e_1$'s classification. Its value $1$ for  $F_2$ is a counterfactual value-explanation, and its explanatory responsibility is
$\mbox{\sf x-Resp}_{\e_1,F_2}(1):=1$. \red{The empty set is its contingency set. Now, entity $\e_4$ shows that the intervened value for $F_2$, i.e. $1$, needs $\{\langle F_1,0\rangle\}$ as contingency set  for the label to switch to $0$.}\boxtheorem
\end{example}

\red{In this work we are interested mostly in c-explanations, which are our {\em best explanations},  and in {\em maximum-responsibility} feature values. Notice that maximum $\mbox{\sf x-Resp}$ scores  can be obtained by concentrating only on c-explanations. \ Maximum-responsibility value-explanations appear in c-explanations, and only in them. C.f. Section \ref{sec:disc} for considerations on s-explanations and features with non-maximum responsibility scores.}

\section{Complexity of \ {\sf x-Resp} Computation}\label{sec:compl}

A natural question is about the complexity of computing the {\sf x-Resp} score. Not only for the obvious reason of knowing the complexity, but also to determine if the ASP-based approach we will present is justified from the complexity point of view. We can shed some light on this matter by taking advantage of complexity results for actual causality in databases \cite{suciu,tocs}. It is known that there are Boolean Conjunctive Queries (BCQs), $\mc{Q}$, for which deciding if the responsibility of a tuple for $\mc{Q}$ being true in a database $D$ is above a given threshold is $\nit{N\!P}$-complete, in the size of $D$ \cite{tocs}.

In our case, given a fixed classifier $\mc{C}$, the computational problem is about deciding, for an arbitrary entity $\e^\star$ and rational number $v$, if $\mbox{\sf x-Resp}(\e^\star) > v$. The complexity is measured as a function of $|\e^\star|$, the size $M$ of cartesian product of the the underlying domains, and the complexity of computing $\mc{C}$. (Under our ASP-based approach, $M$ will be associated to the extensional database for the program.)

Given a relational database schema $\mc{S} = \{R_1, \ldots, R_k\}$, with predicates with arities $\nit{ar}(R_i)$, we can see each attribute $A$ as a feature  that takes values in a finite domain $\nit{Dom}(A)$. Without loss of generality, we assume the predicates $R_i$ do not share attributes (but attributes  may share a same domain). \ Then, we have a sequence of attributes $\langle A^1_1,$ $ \ldots, A^1_{\nit{ar}(R_1)}, \ldots, A^k_1, \ldots, A^k_{\nit{ar}(R_k)}\rangle$. \
For a database $D$, for each of the potential database tuples $\tau_1, \ldots \tau_M$ (in $D$ or not), with $M = |\nit{Dom}(A^1_1)| \times \cdots \times |\nit{Dom}(A^1_{\nit{ar}(R_1)})| + \cdots + |\nit{Dom}(A^k_1)| \times \cdots \times |\nit{Dom}(A^k_{\nit{ar}(A_k)})|$, define binary features $F_j(\tau_j) := 1$ if $\tau_j \in D$, and $0$, otherwise.

Now, define a binary entity $\e^\star = \langle F(\tau_j)\rangle_{\tau_j \in D}$. This entity, containing only $1$s, represents the contents of database $D$, and its length coincides with the number of tuples in $D$, say $|D|$. Now, the population of entities is $\mc{E} = \{1,0\}^{|D|}$, i.e. all the binary sequences with the same length as $\e^\star$. Intuitively, each entity in $\mc{E}$ represents a sub-instance $D^{\e}$ of $D$, by keeping only the tuples of $D$ that are assigned the value $1$. We do not need to consider the insertion of tuples in $D$, as will be clear soon.

Now, given the fixed BCQ $\mc{Q}$, for which $D \models \mc{Q}$, define a binary classifier $\mc{C}^{\mc{Q}}$ as follows: \ For $\e \in E$,  $\mc{C}^{\mc{Q}}(\e) :=1 \mbox{ iff } D^\e \models \mc{Q}$. Notice that this classifier runs in polynomial-time in the  length of $\e$. Also, $\mc{C}^{\mc{Q}}(\e^{\star}) = 1$. The monotonicity of $\mc{Q}$ allows us to concentrate on sub-instances of $D$. Only subinstances can invalidate the query, and superinstances will always make it true.  In this way, we have reduced the problem of computing the responsibility of a tuple $\tau_j$ as an actual cause for $D \models \mc{Q}$ to the problem of computing $\mbox{\sf x-Resp}_{\e^{\star},F_j}\!(1)$.

\begin{theorem} \label{thm:compl} There is a binary polynomial-time classifier $\mc{C}$ over a finite set of binary entities $\mc{E}$ for which deciding if $\mbox{\sf x-Resp}_{\e,F}(v)$ is above a certain threshold is $\nit{N\!P}$-complete in the size of $\e$ plus the size of $\mc{E}$. \boxtheorem
\end{theorem}

So as in \cite{tocs}, several other problems in relation to responsibility can be investigated; and it is likely that most (if not all) the results in \cite{tocs} can be used to obtain similar results for the $\mbox{\sf x-Resp}$ score.

\section{Counterfactual Intervention Programs}\label{sec:CIPs}

An answer-set program has a possible-world semantics, given in terms of its {\em stable models}, which are the intended models of the program \cite{GK14}. A program  consists of a set of rules with variables, possibly with negated atoms in a rule body (antecedent) and disjunctions in the head (consequent). This negation is non-monotonic, which is particulary useful for doing commonsense reasoning and specifying persistence of properties. A program has an {\em extensional database} consisting of ground atoms (the facts). In our case, the facts will be related to the entity under classification for whose label we want counterfactual explanations. The program specifies the possible interventions. Final, intervened versions of the original entity, that have their label switched, correspond to different stable models of the program.

 Entities will be represented by means of a predicate with $n+2$ arguments $E(\cdot;\cdots;\cdot)$. The first one holds a record (or entity) id (which may not be needed when dealing with single entities). The next $n$ arguments hold the feature values.\footnote{For performance-related reasons, it might be more convenient to use $n$ 3-ary predicates to represent an entity with an identifier, but the presentation here would be more complicated.} The last argument holds an annotation constant from the set $\{\ori,\doo,\mathbf{\star},\st\}$. Their semantics will be specified below, by the generic program that uses them.

\red{Initially, a record $\e = \langle f_1, \ldots, f_n\rangle$ has not been subject to interventions, and the corresponding entry in predicate $E$ is of the form $E(\e,\bar{f},\ori)$, where $\e$ is (with a bit of abuse of notation) a constant used a an entity identifier, $\bar{f}$ is an abbreviation for $f_1,\ldots,f_n$, i.e. the feature values for entity $\e$;  and the annotation constant ``$\ori$" stands for ``original entity".}

When the classifier gives label $1$ to $\e$, the idea is to start changing feature values, one at a time. The intervened entity becomes then annotated with constant $\doo$ in the last argument.\footnote{The do-operator is common to denote interventions \cite{pearl}. Here, it is just a constant.}  When the resulting intervened entities are classified, we may not have the classifier specified within the program. For this reason, the program uses a special \red{predicate $\mc{C}(\cdots,\cdot)$, where the first arguments take the feature values of an entity under classification, and the last argument returns the binary label. This predicate can be explicitly given as a set of facts (c.f. Example \ref{ex:second}), or can be specified within the program (c.f. Example \ref{ex:tree2}), or can be invoked by the program as an external predicate (c.f. Example \ref{ex:tree3}), much in the spirit of HEX-programs \cite{eiter1,eiter2}.} \ Since the original instance may have to go through several interventions until reaching one that switches the label to $0$, the intermediate entities get the ``transition" annotation $\bstar$.
\ This is achieved by a generic program.

\vspace{3mm}
\subsection{Counterfactual intervention programs}\label{sec:progr}
The generic {\em Counterfactual Intervention Program} (CIP) is as follows:

\begin{itemize}
\item[P1.] The facts of the program are the atoms of the form \ $\nit{Dom}_i(c)$, with $c \in \nit{Dom}_i$, plus the initial entity \red{$E(\e,\bar{f},\ori)$}, with $\bar{f}$ the initial vector of feature values.
\item[P2.] The transition entities are obtained as initial, original entities, or as the result of an intervention. \ Here, $e$ is a variable standing for a record id.
\begin{eqnarray*}
E(e,\bar{x},\mf{\star}) &\longleftarrow& E(e,\bar{x},\ori).\\
E(e,\bar{x},\mf{\star}) &\longleftarrow& E(e,\bar{x},\doo).
\end{eqnarray*}
\item[P3.] The program rule specifying that, every time the entity at hand (original or obtained after a ``previous" intervention) is classified with label $1$,  a new value has to be picked from a domain, and replaced for the current value.  The new value is chosen via the non-deterministic ``choice operator", well-known in ASP \cite{zaniolo}. In this case, the values are chosen from the domains, subject to the condition of not being the same as the current value: 
    \end{itemize}
\begin{eqnarray*}
\ignore{\mbox{\phantom{ooo}}}E(e,x_1',x_2, \ldots,x_n,\mbox{\bf \sf do}) \vee \cdots \vee E(e,x_1,x_2, \ldots, x_n',\mbox{\bf \sf do})  \ \longleftarrow \ E(e,\bar{x},\mf{\star}), \mc{C}(\bar{x},1),\\ \hspace*{4cm} \nit{Dom}_1(x_1'), \ldots, \nit{Dom}_n(x_n'),x_1'\neq x_1, \ldots, x_n'\neq x_n,\\ \hspace*{4cm}  \nit{choice}(\bar{x},x_1'), \ldots, \nit{choice}(\bar{x},x_n').
\end{eqnarray*}
In general, for each fixed $\bar{x}$, $\nit{choice}(\bar{x},y)$ chooses a unique value $y$ subject to the other conditions in the same rule body. The use of the choice operator can be eliminated by replacing each $\nit{choice}(\bar{x},x_i')$ atom by the atom $\nit{C\!hosen}_i(\bar{x},x_i')$, and defining each predicate $\nit{C\!hosen}_i$ by  means of ``classical" rules \cite{zaniolo}, as follows:\footnote{\red{We emphasize that we are using here the ``choice operator", which is definable in ASP (as done here), and not the newer {\em choice rules}, which could be used here for the same purpose (and many more) and are included in the {\em ASP-Core-2 Standard} \cite{stan}. We use the choice operator, because most of our programs are being run with {\em DLV-Complex}, which does not support choice rules.}}
\begin{eqnarray}
\nit{C\!hosen}_i(\bar{x},y)&\leftarrow&E(e,\bar{x},\mf{\star}),  \ \mc{C}(\bar{x},1), \ \nit{Dom}_i(y), \ y \neq x_i, \nonumber\\&& \mbox{\phantom{oooooooooooo}}\nit{not} \ \nit{Di\!f\!\!f\!Choice}_i(\bar{x},y).  \label{eq:uan}\\
\nit{Di\!f\!\!f\!Choice}_i(\bar{x}, y)&\leftarrow&\nit{C\!hosen}_i(\bar{x}, y'), \ y' \neq y. \label{eq:tu}
\end{eqnarray}

\begin{itemize}
\item[P4.] The following rule specifies that we can ``stop", hence annotation $\st$, when we reach an entity that gets label $0$:\vspace{-2mm}
$$E(e,\bar{x},\mbox{\bf \sf s}) \ \longleftarrow \ E(e,\bar{x},\doo), \ \mc{C}(\bar{x},0).$$

\item[P5.] We add a {em program constraint} specifying that we prohibit going back to the original entity via local interventions:

\vspace{2mm}\hspace*{3cm}$\longleftarrow \ E(e,\bar{x},\doo), \ E(e,\bar{x},\ori).$

\item[P6.]  The counterfactual explanations can be collected by means of a predicate $\nit{Expl}(\cdot,\cdot,\cdot)$ specified by means of:

\vspace{2mm}\hspace*{1cm} $\nit{Expl}(e,i,x_i) \ \longleftarrow \ E(e,x_1,\ldots,x_n,\ori), \ E(e,x_1',\ldots,x_n',\st), \ x_i \neq x_i'$,

\vspace{1mm}with $i = 1, \ldots,n$. \ They collect each value that has been changed in the original instance $e$, with its position in $e$ (the second argument of the predicate). Actually, each of these is a value-explanation.
\boxtheorem
\end{itemize}

The program will have several stable models due to the disjunctive rule and the choice operator.  Each model will hold intervened versions of the original entity, and hopefully versions for which the label is switched, i.e. those with annotation $\st$. If the classifier never switches the label, despite the fact that local interventions are not restricted, we will not find a model with a version of the initial entity annotated with $\st$. \ Due to the constraint in P5., none of the models will have the original entity annotated with $\doo$, because those models would be discarded \cite{leone}.
The definition of the choice operator contains non-stratified negation. The semantics of ASP, which involves model minimality, makes only one of the atoms in a head disjunction true (unless forced otherwise by the program).

\begin{example} (example \ref{ex:first} continued) \label{ex:second} Most of the CIP above is generic. Here we have the facts: \ $\nit{Dom}_1(0),$  $\nit{Dom}_1(1),$ $\nit{Dom}_2(0),  \nit{Dom}_2(1), \nit{Dom}_3(0),$  $\nit{Dom}_3(1)$ and $E(\e_1,0,1,1,\ori)$, with $\e_1$ a constant, the record id of the first row in Table 1.
\ The classifier is explicitly given by Table 1. Then, predicate $\mc{C}$ can be  specified with a set of additional facts:
$\mc{C}(0,1,1,1)$, $\mc{C}(1,1,1, 1)$,
$\mc{C}(1, 1, 0,1)$
$\mc{C}(1, 0, 1,0)$
$\mc{C}(1, 0, 0,1)$
$\mc{C}(0, 1, 0,1)$
$\mc{C}(0, 0, 1,0)$
$\mc{C}(0, 0, 0,0)$. In them, the last entry corresponds the label assigned to the entity whose feature values are given in  the first three arguments.

The stable models of the program will contain all the facts above. One of them, say $\mc{M}_1$, will contain (among others) the facts: \ $E(\e_1,0,1,1;\ori)$ and $E(\e_1,0,1,1;\bstar)$. \ The presence of the last atom activates rule P3., because  $\mc{C}(0,1,1,1)$ is true (for $\e_1$ in Table 1). New facts are produced for $\mc{M}_1$  (the new value due to an intervention is underlined): $E(\e_1,\ul{1},1,1,\doo),$ $ E(\e_1,\ul{1},1,1,\bstar)$. \ Due to the last fact and the true $\mc{C}(1, 1, 1,1)$,
rule P3. is activated again. Choosing the value $0$ for the second disjunct, atoms
$ E(\e_1,\ul{1},\ul{0},1,\doo),$ $ E(\e_1,\ul{1},\ul{0},1,\bstar)$ are generated.  For the latter, $\mc{C}(1, 0, 1,0)$ is true (coming from $\e_4$ in Table 1), switching the label to $0$. Rule P3. is no longer activated, and we can apply rule P4., obtaining \ $E(\e_1,\ul{1},\ul{0},1,\st)$.

 From rule P6., we obtain explanations  $\nit{Expl}(\e_1,1,0), \nit{Expl}(\e_1,2,1)$, showing the changed values in $\e_1$. All this in model $\mc{M}_1$. \ There are other models, and one of them contains $E(\e_1,0,\ul{0},1,\st)$, the minimally intervened version of $\e_1$, i.e. $\e_7$. \boxtheorem
 \end{example}

\red{In the next example we will show how  to write and run the counterfactual intervention program for Example \ref{ex:first} with  the {\em DLV} system \cite{leone}.\footnote{\newred{We have experimented, with the examples in this paper and others,  with each of \href{http://www.dlvsystem.com/?s=manual}{\underline{DLV}},  \href{https://www.mat.unical.it/dlv-complex}{\underline{DLV-Complex}}, and  \href{https://dlv.demacs.unical.it/}{\underline{DLV2}}. They have been extremely useful. At this moment, each of them seems to have some nice features the others lack.}} When numerical aggregations and, specially, set operations are needed, we use instead the {\em DLV-Complex} system \cite{complex} (c.f. Section \ref{sec:disc}). In Example \ref{ex:tree3} we show a program that can be used with the newer version, {\em DLV2} \cite{dlv2,wasp}, of {\em DLV}, which follows the ASP-Core-2 standard \cite{stan}.   }

 \begin{example} \label{ex:dlv} (example \ref{ex:second} continued) \red{The answer-set program for Examples \ref{ex:first} and \ref{ex:second}, written in the language for the {\em DLV-Complex} system is shown next. (The program portion shown right below would also run with {\em DLV} since it does not contain numerical aggregations.)} In it, the annotation ``\verb+tr+" stands for the transition annotation ``$\star$" used in Example \ref{ex:second}, and \verb+X, Xp+ stand for $x, x^\prime$, etc. In a {\em DLV} program, terms starting with a lower-case letter are constants; and those starting with an upper-case letter are variables.

{\footnotesize \begin{verbatim}
    #include<ListAndSet>
    % the classifier:
    cls(0,1,1,1). cls(1,1,1,1). cls(1,1,0,1). cls(1,0,1,0). cls(1,0,0,1).
    cls(0,1,0,1). cls(0,0,1,0). cls(0,0,0,0).
    % the domains:
    dom1(0). dom1(1). dom2(0). dom2(1). dom3(0). dom3(1).
    % original entity at hand:
    ent(e,0,1,1,o).

    % transition rules:
    ent(E,X,Y,Z,tr) :- ent(E,X,Y,Z,o).
    ent(E,X,Y,Z,tr) :- ent(E,X,Y,Z,do).

    % admissible counterfactual interventions:
    ent(E,Xp,Y,Z,do) v ent(E,X,Yp,Z,do) v ent(E,X,Y,Zp,do) :- ent(E,X,Y,Z,tr),
                                    cls(X,Y,Z,1), dom1(Xp), dom2(Yp), dom3(Zp),
                                    X != Xp, Y != Yp, Z!= Zp, chosen1(X,Y,Z,Xp),
                                    chosen2(X,Y,Z,Yp), chosen3(X,Y,Z,Zp).

    % definitions of chosen operators as in equations (2) and (3):
    chosen1(X,Y,Z,U) :- ent(E,X,Y,Z,tr), cls(X,Y,Z,1), dom1(U), U != X,
                        not diffchoice1(X,Y,Z,U).
    diffchoice1(X,Y,Z, U) :- chosen1(X,Y,Z,Up), U != Up, dom1(U).
    chosen2(X,Y,Z,U) :- ent(E,X,Y,Z,tr), cls(X,Y,Z,1), dom2(U), U != Y,
                        not diffchoice2(X,Y,Z,U).
    diffchoice2(X,Y,Z, U) :- chosen2(X,Y,Z,Up), U != Up, dom2(U).
    chosen3(X,Y,Z,U) :- ent(E,X,Y,Z,tr), cls(X,Y,Z,1), dom3(U), U != Z,
                        not diffchoice3(X,Y,Z,U).
    diffchoice3(X,Y,Z, U) :- chosen3(X,Y,Z,Up), U != Up, dom3(U).

    % stop when label has been changed:
    ent(E,X,Y,Z,s) :- ent(E,X,Y,Z,do), cls(X,Y,Z,0).

    % hard constraint for not returning to original entity:
    :- ent(E,X,Y,Z,do), ent(E,X,Y,Z,o).

    % auxiliary predicate to avoid unsafe negation in the hard constraint below:
    entAux(E) :- ent(E,X,Y,Z,s).

    % hard constraint for not computing models where label does not change:
    :- ent(E,X,Y,Z,o), not entAux(E).

    % collecting explanatory changes per argument:
    expl(E,1,X) :- ent(E,X,Y,Z,o), ent(E,Xp,Yp,Zp,s), X != Xp.
    expl(E,2,Y) :- ent(E,X,Y,Z,o), ent(E,Xp,Yp,Zp,s), Y != Yp.
    expl(E,3,Z) :- ent(E,X,Y,Z,o), ent(E,Xp,Yp,Zp,s), Z != Zp.
 \end{verbatim} }

 If we run this program with DLV-Complex, we obtain the following three models; for which we do not show (most of) the original program facts, as can be requested from DLV:
 {\footnotesize \begin{verbatim}

    {ent(e,0,1,1,o), ent(e,0,1,1,tr), chosen1(0,1,1,1), chosen2(0,1,1,0),
     chosen3(0,1,1,0), ent(e,0,0,1,do), ent(e,0,0,1,tr), ent(e,0,0,1,s),
     diffchoice3(0,1,1,1), diffchoice2(0,1,1,1), diffchoice1(0,1,1,0),
     entAux(e), expl(e,2,1)}

    {ent(e,0,1,1,o), ent(e,0,1,1,tr), chosen1(0,1,1,1), chosen2(0,1,1,0),
     chosen3(0,1,1,0), ent(e,0,1,0,do), ent(e,0,1,0,tr), chosen1(0,1,0,1),
     chosen2(0,1,0,0), chosen3(0,1,0,1), ent(e,0,0,0,do), ent(e,0,0,0,tr),
     ent(e,0,0,0,s), diffchoice3(0,1,0,0), diffchoice3(0,1,1,1),
     diffchoice2(0,1,0,1), diffchoice2(0,1,1,1), diffchoice1(0,1,0,0),
     diffchoice1(0,1,1,0), entAux(e), expl(e,2,1), expl(e,3,1)}

    {ent(e,0,1,1,o), ent(e,0,1,1,tr), chosen1(0,1,1,1), chosen2(0,1,1,0),
     chosen3(0,1,1,0), ent(e,1,1,1,do), ent(e,1,1,1,tr), chosen1(1,1,1,0),
     chosen2(1,1,1,0), chosen3(1,1,1,0), ent(e,1,0,1,do), ent(e,1,0,1,tr),
     ent(e,1,0,1,s), diffchoice3(0,1,1,1), diffchoice3(1,1,1,1),
     diffchoice2(0,1,1,1), diffchoice2(1,1,1,1), diffchoice1(0,1,1,0),
     diffchoice1(1,1,1,1), entAux(e), expl(e,1,0), expl(e,2,1)}
 \end{verbatim}  }

\vspace{-3mm}\noindent These models correspond to the counterfactual entities $\e_7, \e_8, \e_4$, resp. in Example \ref{ex:first}.

\red{Notice that the program, except for the fact} \verb+ent(e,0,1,1,o)+ \red{in the 10th line, is completely generic, and can be used with any input entity that has been classified with label $1$.}\footnote{If the initial label is $0$ instead, no interventions would be triggered, and the only model would correspond to the initial entity.} We could remove it from the program, obtaining
program \verb+theProgram2.txt+, and we could run instead

\hspace*{1.2cm} \verb+C:\DLV>dlv.exe ent.txt program2.txt > outputFile.txt+

\noindent where \verb+ent.txt+ is the file containing only \ \verb+ent(e,0,1,1,o).+ .
 \boxtheorem
 \end{example}

 \red{In the previous example, the classifier was given as an input/output relation, that is, as a set of facts inserted directly in the program. In other situations, we may have the classifier invoked from the program as an external predicate. In others, the classifier can be specified directly in the program, as shown in Example \ref{ex:tree}.}

\red{Our CIPs compute all the {\em counterfactual versions} (or counterfactual explanations) of the original entity $\e$. Each counterfactual version is represented (or characterized) by at least one of the stable models of the CIP; one that contains the counterfactual version of $\e$ annotated with ``$\msf{s}$". There may be more that one stable model associated to a counterfactual version $\e'$ due to the use of the choice operator. Different choices may end up leading to the same $\e'$. }

\red{The counterfactual explanations obtained through the CIP are not necessarily s-explanations or c-explanations (c.f. Section \ref{sec:causes}), as Example \ref{ex:dlv} shows. The CIPs presented so far have (only) minimal models with respect to set-inclusion of the extensions of full predicates, whereas when we compare explanations, we do it at the ``attribute (or feature, or argument) level". Of course, s-explanations and c-explanations are all included among the counterfactual explanations, and are represented by stable models of the CIP.  We will consider this point in Section \ref{sec:cexp}.}

\begin{example} \label{ex:nonMin} (example \ref{ex:first} continued) \newred{To show the different kinds of counterfactual versions of an original entity, let us change the classifier we had before by the one shown in Table 2.}

\begin{multicols}{2}
\begin{center}
\begin{tabular*}{4cm}{|c|| c|c|c||c|}\cline{1-5}
entity (id) & $F_1$ & $F_2$ & $F_3$ & $L$\\ \cline{1-5}
$\e_1^\prime$ & 0 & 1 & 1 &1\\ \cline{1-5}
$\e_2^\prime$ & 1 & 1 & 1 &1\\
$\e_3^\prime$ & \underline{1} & 1 & \underline{0} &0\\
$\e_4^\prime$ & 1 & 0 & 1 &1\\
$\e_5^\prime$ & \underline{1} & \underline{0} & \underline{0} &0\\
$\e_6^\prime$ & 0 & 1 & 0 &1\\
$\e_7^\prime$ & 0 & \underline{0} & 1 &0\\
$\e_8^\prime$ & 0 & 0 & 0 &1\\ \cline{1-5}
\end{tabular*}\\ \vspace{2mm}
{\bf Table 2: \ Classifier \ $\mc{C}^\prime$}
\end{center}


The changed values appear underlined.  \
In this case, we have three counterfactual versions: (a) $\e_7^\prime$ that is c-explanation. Only one value is changed to switch the label to $0$. (b) $\e_3^\prime$ is an s-explanation, but not a c-explanation. It shows two changes, but not the one for $\e_7^\prime$. (c) $\e_5^\prime$ is neither an s- nor a c-explanation. Its changes include those for $\e_7^\prime$ and those for  $\e_3^\prime$.
\end{multicols}

 The CIP we have so far would return the three of them. \
With the additional elements for CIP programs to be introduced in Section \ref{sec:cexp}, we will be able to obtain only c-explanations. In Section \ref{sec:disc}, we briefly discuss the case of s-explanations (that may not be c-explanations). \boxtheorem
\end{example}

\subsection{Complexity of CIPs}\label{sec:hcf}

The complexity result  obtained in Section \ref{sec:compl} makes us wonder whether using ASP-programs for specifying and computing the $\mbox{\sf x-Resp}$ score is an overkill, or, more precisely,
whether  they have the right and necessary expressive power and complexity  to confront our problem. In fact they do. It is known that reasoning with disjunctive answer-set programs (DASP) falls in the second level of the {\em polynomial hierarchy} (in data complexity) \cite{dantsin}, and slightly above that by the use of {\em weak constraints} \cite{leone} that we will use in Section \ref{sec:cexp}. However, CIPs have the  property of being {\em head-cycle free} (HCF), which brings down the complexity of a DASP to the first level of the polynomial hierarchy \cite{dantsin}. \red{This is in line with the result in Theorem \ref{thm:compl}.}

\ignore{In order to see that this HCF property holds, we introduce the {\em dependency graph} of a ground DASP  $\Pi$ as a
directed graph whose nodes are the atoms in $\Pi$, with an edge from $L$ to $L'$
iff there is a rule in which $L$ appears positive in the body and $L'$ appears in
the head. $\Pi$ is HCF iff its dependency graph does not contain
directed cycles  through two atoms that belong to the head of the same
rule.}

\red{It is easy to check  that CIPs are HCF (c.f. Section \ref{sec:backgr}): The ground chains in the  directed graph $\nit{DG}(\Pi)$ associated to a CIP $\Pi$ are, due to rules P2. and P3.,  of the form: \ $E(\e,\bar{a}, \doo) \rightarrow E(\e,\bar{a}, \bstar) \rightarrow E(\e,\bar{a}', \doo)$, with $\bar{a} \neq \bar{a}'$. They never create a cycle in the head of a ground instantiation of the disjunctive rule.}

\red{One can also see the HCF property from the fact that the CIPs become {\em repair-programs} \cite{bertossi11} for a database w.r.t. the integrity constraint, actually \ignore{ $\forall e \forall \bar{x}(E(e,\bar{x}) \wedge \mc{C}(\bar{x},1) \ \rightarrow \ \bot)$, with $\bot$ an always false propositional atom. This constraint is equivalent to}  {\em denial constraint}, \ $\forall e \forall \bar{x} \neg (E(e,\bar{x}) \wedge \mc{C}(\bar{x},1))$.
Denial constraints are common in databases, and their repairs and repair-programs have been investigated  \cite{monica,foiks18}. C.f. \cite{bertossi11} for additional references.}
\ignore{ Repairs programs for classes of ICs of the form $\bar{x}(C(\bar{x}) \rightarrow \psi)$, where $C(\bar{x})$ is a conjunction of atoms, and $\psi$ contains only built-ins, are always HCF \cite{barcelo,monica}. }

\red{As a consequence of being HCF, a CIP can be transformed, by means of the {\em shift operation}, into an equivalent non-disjunctive ASP (c.f. Section \ref{sec:backgr})}. \ignore{ by moving in turns all the head atoms but one in negated form to the body (and so generating as many non-disjunctive rules as atoms in the head of the original disjunction) \cite{Ben94,dantsin}.}

\begin{example} (example \ref{ex:second} continued) The disjunctive rule in Example \ref{ex:second} can be replaced by the three rules:
{\footnotesize \begin{verbatim}
 ent(E,Xp,Y,Z,do) :- ent(E,X,Y,Z,tr), cls(X,Y,Z,1), dom1(Xp), dom2(Yp), dom3(Zp),
                     X != Xp, Y != Yp, Z!= Zp, chosen1(X,Y,Z,Xp), chosen2(X,Y,Z,Yp),
                     chosen3(X,Y,Z,Zp), not ent(E,X,Yp,Z,do), not ent(E,X,Y,Zp,do).

 ent(E,X,Yp,Z,do) :- ent(E,X,Y,Z,tr), cls(X,Y,Z,1), dom1(Xp), dom2(Yp), dom3(Zp),
                     X != Xp, Y != Yp, Z!= Zp, chosen1(X,Y,Z,Xp), chosen2(X,Y,Z,Yp),
                     chosen3(X,Y,Z,Zp), not ent(E,Xp,Y,Z,do), not ent(E,X,Y,Zp,do).

 ent(E,X,Y,Zp,do) :- ent(E,X,Y,Z,tr), cls(X,Y,Z,1), dom1(Xp), dom2(Yp), dom3(Zp),
                     X != Xp, Y != Yp, Z!= Zp, chosen1(X,Y,Z,Xp), chosen2(X,Y,Z,Yp),
                     chosen3(X,Y,Z,Zp), not ent(E,Xp,Y,Z,do), not ent(E,X,Yp,Z,do).
\end{verbatim}}
The resulting program has the same answer-sets as the original program. \boxtheorem
\end{example}

\subsection{C-explanations and maximum responsibility}\label{sec:cexp}

\red{As discussed at the end of Section \ref{sec:progr}, an intervened entity of the form $E(\e,c_1,\ldots,$ $ c_n,\st)$, that is, representing a counterfactual explanation, may not correspond to an s- or a c-explanation. We are interested in obtaining the latter,  and only them, because: (a) They are our ``best explanations", and (b) They are used to define and compute the {\em maximum} $\mbox{\sf x-Resp}$ scores. }

\red{Moving towards computing $\mbox{\sf x-Resp}$ scores, notice that in each of the stable models $\mc{M}$ of the CIP, we can collect the corresponding  counterfactual explanation for $\e$'s classification as the set \ $\epsi^{\mc M} = \{\langle F_i, c_i\rangle~|$ $\nit{Expl}(\e,i;c_i)$  $\in \mc{M}\}$. This can be done within a ASP system such as {\em DLV}, which allows set construction and aggregation, in particular, counting \cite{leone}. Actually, counting comes handy to obtain the cardinality of  $\epsi^{\mc M}$, by means of:
\begin{equation}
\nit{inv\mbox{-}resp}(\e,m) \longleftarrow \#count\{i: \nit{Expl}(\e,i;c_i)\}=m.\label{eq:m}
\end{equation}
For each model $\mc{M}$ of the CIP,  we will obtain such a value $m(\mc{M})$ that shows the number of changes of feature values that lead to the associated counterfactual explanation. Notice that, in each of the models $\mc{M}^{\!o}$ that correspond to c-explanations, these, now minimum values $m(\mc{M}^{\!o})$ will be the same, say $m^{\!o}$, and can be used to compute the responsibility of a feature value in $\epsi^{{\mc M}^o}$, as follows: For  $\nit{Expl}(\e,i;c_i) \in \mc{M}^{\!o}$,
\begin{equation}
\mbox{\sf x-Resp}_{\e,F_i}(c_i) = \frac{1}{|\epsi^{\mc M}|} = \frac{1}{m^{\!o}}. \label{eq:resp} 
\end{equation}}

\begin{example} (example \ref{ex:dlv} continued) \label{ex:resp} \red{Let us add to the CIP above the rule:}

{\footnotesize \begin{verbatim}
    % computing the inverse of x-Resp:
    invResp(E,M) :- #count{I: expl(E,I,_)} = M, #int(M), E = e.
\end{verbatim}
 }

\vspace{-1mm} \noindent By running DLV-Complex with the new program, we obtain the models above extended with atoms representing the changes in arguments of the original entity (we omit most of the old atoms):

\vspace{-1mm}
 {\footnotesize \begin{verbatim}

    {ent(e,0,1,1,o), ... , ent(e,0,0,1,s), expl(e,2,1), invResp(e,1)}
    {ent(e,0,1,1,o), ..., ent(e,0,0,0,s), expl(e,2,1), expl(e,3,1), invResp(e,2)}
    {ent(e,0,1,1,o), ..., ent(e,1,0,1,s), expl(e,1,0), expl(e,2,1), invResp(e,2)}
 \end{verbatim}  }

\vspace{-3mm}\red{As before, we obtain three models, and each of them shows, in the last atom, the number of changes that were made to obtain the corresponding counterfactual explanation.
\ For example, for the last model, say $\mc{M}_3$, corresponding to entity $\e_4$, we obtain $m(\mc{M}_3) = 2$. Similarly for the second one, corresponding to entity $\e_8$. The first model corresponds to the counterfactual entity $\e_7$ that is a c-explanation, which is shown by the minimum value ``$1$" that predicate } \verb+inResp+ takes in its second argument among all the stable models.

 We can see that the first model is the one corresponding to a maximum responsibility feature value.
\boxtheorem
\end{example}

\red{In order to obtain only the models associated to c-explanations,} we
add {\em weak program constraints} to the CIP. They  can be violated by a stable model of the program (unlike strong program constraints), but the {\em number} of violations has to be minimized. In this case, for $1 \leq i \leq n$, we add to the CIP:\footnote{This notation follows the standard in \cite{stan,dlv2}.}
\begin{equation}
:\sim \ E(e,x_1,\ldots,x_n,\ori), \ E(e,x_1',\ldots,x_n',\st), x_i \neq x_i'.\label{eq:weak}
\end{equation}
Only the stable models representing an intervened version of $\e$ with a minimum number of value discrepancies with $\e$ will be kept.

\begin{example}  (example \ref{ex:resp} continued)
 The best explanations, i.e. the c-explanations,  can be obtained by adding weak program constraints to the combined CIP above:

\vspace{-1mm}
 {\footnotesize \begin{verbatim}
    % weak constraints for minimizing number of changes:
    :~ ent(E,X,Y,Z,o), ent(E,Xp,Yp,Zp,s), X != Xp.
    :~ ent(E,X,Y,Z,o), ent(E,Xp,Yp,Zp,s), Y != Yp.
    :~ ent(E,X,Y,Z,o), ent(E,Xp,Yp,Zp,s), Z != Zp.
  \end{verbatim}  }

 \vspace{-3mm} If we run DLV-Complex with the extended program, we obtain a single model, corresponding to $\e_7$:

\vspace{-1mm}
   {\footnotesize \begin{verbatim}
   Best model: {ent(e,0,1,1,o), ent(e,0,1,1,tr), chosen1(0,1,1,1), chosen2(0,1,1,0),
                chosen3(0,1,1,0), ent(e,0,0,1,do), ent(e,0,0,1,tr), ent(e,0,0,1,s),
                diffchoice3(0,1,1,1), diffchoice2(0,1,1,1), diffchoice1(0,1,1,0),
                expl(e,2,1), entAux(e), invResp(e,1)}
   \end{verbatim}  }

   \red{ This model shows that counterfactual entity $\e_7$ has one change in the second attribute wrt. the original entity. This new entity gives  a minimum  inverse responsibility $m^{\!o} = 1$ to the original value in the second argument of} \verb+ent+, which leads, via (\ref{eq:resp}), to its (maximum) responsibility $\mbox{\sf x-Resp}_{\e_1,F_2}(1)= \frac{1}{m^{\!o}} = 1$. \boxtheorem
\end{example}

\vspace{-4mm}\begin{example} \label{ex:tree2} (example \ref{ex:tree} continued) \ We present now the CIP for the classifier based on the decision-tree, in {\em DLV-Complex} notation. Notice that after the facts, that now do not include the classifier, we find the rule-based specification of the decision tree.

\vspace{-3mm}{\footnotesize
\begin{verbatim}

 #include<ListAndSet>

 % facts:
    dom1(sunny). dom1(overcast). dom1(rain). dom2(high). dom2(normal).
    dom3(strong). dom3(weak).
    ent(e,sunny,normal,weak,o).   % original entity at hand

 % spec of the classifier:
    cls(X,Y,Z,1) :- Y = normal, X = sunny, dom1(X), dom3(Z).
    cls(X,Y,Z,1) :- X = overcast, dom2(Y), dom3(Z).
    cls(X,Y,Z,1) :- Z = weak, X = rain, dom2(Y).
    cls(X,Y,Z,0) :- dom1(X), dom2(Y), dom3(Z), not cls(X,Y,Z,1).

 % transition rules:
    ent(E,X,Y,Z,tr) :- ent(E,X,Y,Z,o).
    ent(E,X,Y,Z,tr) :- ent(E,X,Y,Z,do).

 % counterfactual rule
    ent(E,Xp,Y,Z,do) v ent(E,X,Yp,Z,do) v ent(E,X,Y,Zp,do) :-
                            ent(E,X,Y,Z,tr), cls(X,Y,Z,1), dom1(Xp), dom2(Yp),
                            dom3(Zp), X != Xp, Y != Yp, Z!= Zp,
                     chosen1(X,Y,Z,Xp), chosen2(X,Y,Z,Yp), chosen3(X,Y,Z,Zp).

 % definitions of chosen operators:
    chosen1(X,Y,Z,U) :- ent(E,X,Y,Z,tr), cls(X,Y,Z,1), dom1(U), U != X,
                        not diffchoice1(X,Y,Z,U).
    diffchoice1(X,Y,Z, U) :- chosen1(X,Y,Z, Up), U != Up, dom1(U).
    chosen2(X,Y,Z,U) :- ent(E,X,Y,Z,tr), cls(X,Y,Z,1), dom2(U), U != Y,
                        not diffchoice2(X,Y,Z,U).
    diffchoice2(X,Y,Z, U) :- chosen2(X,Y,Z, Up), U != Up, dom2(U).
    chosen3(X,Y,Z,U) :- ent(E,X,Y,Z,tr), cls(X,Y,Z,1), dom3(U), U != Z,
                        not diffchoice3(X,Y,Z,U).
    diffchoice3(X,Y,Z, U) :- chosen3(X,Y,Z, Up), U != Up, dom3(U).

 % Not going back to initial entity (program constraint):
    :- ent(E,X,Y,Z,do), ent(E,X,Y,Z,o).

 % stop when label has been changed:
    ent(E,X,Y,Z,s) :- ent(E,X,Y,Z,do), cls(X,Y,Z,0).

 % collecting changed values for each feature:
    expl(E,outlook,X)   :- ent(E,X,Y,Z,o), ent(E,Xp,Yp,Zp,s), X != Xp.
    expl(E,humidity,Y)  :- ent(E,X,Y,Z,o), ent(E,Xp,Yp,Zp,s), Y != Yp.
    expl(E,wind,Z)      :- ent(E,X,Y,Z,o), ent(E,Xp,Yp,Zp,s), Z != Zp.

    entAux(E) :- ent(E,X,Y,Z,s).        % auxiliary predicate to
                                        % avoid unsafe negation
                                        % in the constraint below
    :- ent(E,X,Y,Z,o), not entAux(E).   % discard models where
                                        % label does not change

 % computing the inverse of x-Resp:
    invResp(E,M) :- #count{I: expl(E,I,_)} = M, #int(M), E = e.
\end{verbatim} }

\ignore{++
+++++++++++

\red{Using DLV notation, the CIP  for the decision-tree classifier contains the facts}

{\footnotesize \begin{verbatim}
    dom1(sunny). dom1(overcast). dom1(rain). dom2(high). dom2(normal).
    dom3(strong). dom3(weak). ent(e,sunny,normal,weak,o).
\end{verbatim}
}

\noindent and also a definition of the classifier by means of the rules

{\footnotesize \begin{verbatim}
    cls(X,Y,Z,1) :- Y = normal, X = sunny, dom1(X).
    cls(X,Y,Z,1) :- X = overcast, dom2(Y), dom3(Z).
    cls(X,Y,Z,1) :- Z = weak, X = rain, dom2(Y).
    cls(X,Y,Z,0) :- dom1(X), dom2(Y), dom3(Z), not cls(X,Y,Z,1).
\end{verbatim}
}
The complete CIP is similar to that in Example \ref{ex:dlv}, and can be found  in \ref{sec:app}.++}

 Two counterfactual versions of $\e$ are obtained, as represented by the two essentially different stable models of the program, and determined by the atoms with the  annotation \verb+s+ (we keep in them  only the most relevant atoms, omitting initial facts and choice-related atoms):

 {\footnotesize
\begin{verbatim}
 {ent(e,sunny,normal,weak,o),cls(sunny,normal,strong,1),cls(sunny,normal,weak,1),
  cls(overcast,high,strong,1),cls(overcast,high,weak,1),cls(rain,high,weak,1),
  cls(overcast,normal,weak,1),cls(rain,normal,weak,1),cls(overcast,normal,strong,1),
  cls(sunny,high,strong,0),cls(sunny,high,weak,0),cls(rain,high,strong,0),
  cls(rain,normal,strong,0),ent(e,sunny,high,weak,do),ent(e,sunny,high,weak,tr),
  ent(e,sunny,high,weak,s),expl(e,humidity,normal),invResp(e,1)}

 {ent(e,sunny,normal,weak,o), cls(sunny,normal,strong,1),...,
  cls(rain,normal,strong,0),ent(e,rain,normal,strong,do),ent(e,rain,normal,strong,tr),
  ent(e,rain,normal,strong,s),expl(e,outlook,sunny),expl(e,wind,weak),invResp(e,2)}
\end{verbatim} }

The first model shows the classifiers as a set of atoms, and its last line, that \verb+ent(e,sunny,high,weak,s)+ is a counterfactual version, with label $0$, of the original entity $\e$, and is obtained from
the latter by means of changes of values in feature $\mathsf{Humidity}$, leading to an inverse score of $1$. \ The second model shows a different counterfactual version of $\e$, namely \verb+ent(e,rain,normal,strong,s)+, now obtained by changing values for features $\mathsf{Outlook}$ and $\mathsf{Wind}$, leading to an inverse score of $2$.

Let us now add, at the end of the program the following weak constraints (labeled with \verb+(*)+):

{\footnotesize \begin{verbatim}
 % Weak constraints to minimize number of changes:      (*)
    :~ ent(E,X,Y,Z,o), ent(E,Xp,Yp,Zp,s), X != Xp.
    :~ ent(E,X,Y,Z,o), ent(E,Xp,Yp,Zp,s), Y != Yp.
    :~ ent(E,X,Y,Z,o), ent(E,Xp,Yp,Zp,s), Z != Zp.
\end{verbatim} }
If we run the program with them, the number of changes is minimized, and we basically obtain only the first model above, corresponding to the counterfactual entity
$\e' = \nit{ent}(\mathsf{sunny},\mathsf{high},\mathsf{weak})$. \boxtheorem
\end{example}
As can be seen at the light of this example, more complex rule-based classifiers could be defined inside  a CIP. It is also possible to invoke the classifier as an external predicate, as the following example shows.

\begin{example}\label{ex:tree3} (example \ref{ex:tree2} continued) \red{The program below calls the classifiers through a predicate that has an external extension, as defined by a Python program.}
\ The program has  the same facts and the same rules as the the program in Example \ref{ex:tree2}, except for a new rule that defines the classification predicate, \verb+cls+ here (and $\mc{C}$ in the general formulation in Section \ref{sec:progr}), and replaces the internal specification of the classifier:

{\small
\begin{verbatim}
     cls(X,Y,Z,L) :- &classifier(X,Y,Z;L), dom1(X), dom2(Y), dom3(Z).
\end{verbatim}}

Here, the atom \verb+&classifier(X,Y,Z;L)+ corresponds to the invocation of the external classifier with parameters \verb+X,Y,Z+, which gets an external  value through variable \verb+L+. The program was run with the version of DLV2 for Linux that supports interaction with Python, and can be downloaded from:

\vspace{-1mm}
{\footnotesize \begin{verbatim}
     https://dlv.demacs.unical.it/publications#h.cgg9mbi41jq9
\end{verbatim}}

The program in Python that specifies the classifier is very simple, and it can be invoked in combination with DLV2, as follows:

\verb+ sudo ./dlv2-python-linux-x86_64   program_dlv2.txt   def_class.py+.

\noindent Here, \verb+program_dlv2.txt+ is the CIP, and \verb+def_class.py+ is the very simple Python program that specifies the classifier, namely

{\footnotesize \begin{verbatim}
 def classifier(X,Y,Z):
        if (X == "sunny") and (Y == "normal"):
                return 1
        if (X == "overcast"):
                return 1
        if (X == "rain") and (Z == "weak"):
                return 1
        else:
                return 0
\end{verbatim}  }

We obtain as  answer-set the first one in Example \ref{ex:tree2}. \boxtheorem
\end{example}

\section{Semantic Knowledge}\label{sec:sem}

Counterfactual interventions in the presence of semantic conditions requires consideration. As the following example shows, not every intervention, or combination of them, may be admissible \cite{jdiq}. In these situations declarative approaches to counterfactual interventions become particularly useful.

 \begin{example}  A moving company makes automated hiring decisions based on feature values  in applicants' records of the form  $R=\langle \nit{appCode}, \mbox{\nit{ability to lift}}, \mbox{\it gender},$ $ \mbox{\it weight},\mbox{\it height}, \nit{age}\rangle$. \
 Mary, represented by  $R^\star = \langle 101, 1, F, \mbox{\it 160 pounds}, \mbox{\it 6 feet}, 28\rangle$ applies, but  is denied the job, i.e. the classifier returns: \ $L(R^\star) = 1$. \
  To explain the decision, we can, hypothetically, change Mary's gender, from $\nit{F}$ into $\nit{M}$, obtaining record $R^{\star\prime}$, for which we now observe $L(R^{\star \prime}) = 0$.  Thus, her value $F$ for {\em gender}  can be seen as a counterfactual explanation for the initial decision.

  As an alternative, we might keep the value of \nit{gender}, and  counterfactually change other feature values. However, we might be constrained or guided by an ontology containing, e.g. the denial semantic constraint \  $\neg (R[2] =1 \wedge R[6]  > 80)$  ($2$ and $6$ indicating positions in the record)  that prohibits someone over 80 to be qualified as fit for lifting weight. \ We could also have a rule, such as \ $ (R[3] = M \wedge R[4] > 100 \wedge R[6] < 70)  \rightarrow R[2] =1$, specifying that men who weigh over 100 pounds and are younger than 70 are automatically qualified to lift weight.

In situations like this, we could add to the CIPs we had before: (a) program constraints that prohibit certain models, e.g.$$\longleftarrow \ R(e,x, 1, y, z,u, w,\bstar), \ w>  80;$$
(b) additional rules, e.g.
$$R(e,x, 1, y , z,u, w,\bstar) \longleftarrow R(e,x, y, M, z,u, w,\bstar), \ z > 100, w < 70,$$
 that may automatically generate additional interventions. In a similar way, one could accommodate certain preferences using weak program constraints.  \boxtheorem
\end{example}

 Causality and responsibility in databases in the presence of integrity constraints was introduced and investigated in \cite{flairsExt}.

 \begin{example} (example \ref{ex:tree2} continued) \red{It might be the case that in a particular region, some combinations of weather conditions are never possible, e.g. raining with a strong wind at the same time.} When producing counterfactual interventions for the entity $\e$, such a combination should be prohibited. This can be done by imposing a  hard program constraint, that we add to the program in Example \ref{ex:tree2}:

 \vspace{1mm}
{\footnotesize
\begin{verbatim}
 % hard constraint disallowing a particular combination    (**)
    :- ent(E,rain,X,strong,tr).
 \end{verbatim} }

 \vspace{-3mm}
If we run in {\em DLV-Complex} the program with this constraint, but without the weak constraints labeled with \verb+(*)+ in Example \ref{ex:tree2},  we obtain only the first model shown in Example \ref{ex:tree2}, corresponding to the counterfactual entity $\e' = \nit{ent}(\mathsf{sunny},\mathsf{high},\mathsf{weak})$.
 \boxtheorem
 \end{example}

\red{As the previous example shows, we can easily impose constraints that make the counterfactual entities, or equivalently, the associated explanations,  {\em actionable} \cite{ustun,karimiB}. As mentioned in Section \ref{sec:intro}, for the loan application example, we could impose a hard program constraint (i.e. add it to a CIP) of the  form}
\red{\begin{equation}
\longleftarrow \nit{E}(e, \ldots, \nit{age}, \ldots,\msf{o}), \ \nit{E}(e, \ldots, \nit{age}', \ldots, \msf{*}), \ \nit{age}' < \nit{age},\label{eq:act}
\end{equation}}
\noindent \red{which prevents decreasing an applicant's age. }

\red{Logic-based specifications also allow for {\em compilation of constraints} into rules. For example, instead of using a hard constraint, such as (\ref{eq:act}), we could directly
impose the condition on a counterfactual age in the disjunctive counterfactual rule P3., of the form
\begin{eqnarray*}
 \cdots \vee E(e,x_1,\ldots, x_a', \ldots, \mbox{\bf \sf do}) \vee \cdots \ \longleftarrow \ E(e,\bar{x},\mf{\star}), \mc{C}(\bar{x},1), \ldots, \nit{Dom}_a(x_a'),\\ \ldots, x_a < x_a',  \ldots, \nit{choice}(\bar{x},x_a'), \ldots.
\end{eqnarray*}
Here, the subscript $a$ refers to the domain and variables for the $\mbox{\sf Age}$ feature. On this basis, one could only increase the age. If this intervention leads to a successful counterfactual  entity (i.e. with a positive label), we could  tell the applicant that he/she has to wait to possibly succeed with the loan application.}

\red{We could also think of explicitly specifying {\em actionable} counterfactual entities, starting with a rule of the form
$$ E(e,\bar{x},\mf{a})\  \longleftarrow \ E(e,\bar{x},\mf{s}), \cdots,$$
where the new annotation $\mf{a}$ stands for ``actionable", and the rule defines an entity as actionable if it is a counterfactual (final) entity that satisfies some extra conditions.}

\red{Several possibilities offer themselves in this direction. All of them require simple, symbolic changes in the overall specification. Doing something similar with a purely procedural approach would be much more complex, and would require modifying the underlying code. }

Another situation where not all interventions are admissible occurs when
features take continuous values, and their domains have to be discretized. The common way of doing this, namely the combination of {\em bucketization and one-hot-encoding}, leads to the natural and necessary imposition of additional constraints on interventions, as we will show. \ Through
bucketization, a feature range is discretized by splitting it into finitely many, say $N$, usually non-overlapping intervals. This makes the feature basically categorical (each interval becoming a categorical value). Next, through one-hot-encoding, the original feature is represented as a vector of length $N$ of indicator functions, one for each categorical value (intervals here) \cite{deem}. In this way, the original feature gives rise to $N$ binary features. For example, if we have a continuous feature ``External Risk Estimate" ($\mbox{\sf ERE}$), its buckets could be: $[0, 64), [64, 71), [71, 76), [76, 81),$ $ [81,\infty)$. Accordingly, if
for an entity $\e$,  $\mbox{\sf ERE}(\e) = 65$, then, after one-hot-encoding, this value is represented as the vector $[0, 1, 0, 0, 0, 0]$, because $65$ falls into the second bucket.

In a case like this, it is clear that counterfactual interventions are constrained by the assumptions behind  bucketization and one-hot-encoding. For example, the vector cannot be updated into, say $[0, 1, 0, 1, 0, 0]$, meaning that the feature value for the entity falls both in intervals $[64, 71)$ and $[76, 81)$.
\ Bucketization and one-hot-encoding can make  use of program constraints, such as \ $\longleftarrow \mbox{\sf ERE}(e,x,1,y,1,z,w,\bstar)$, etc. \ Of course, admissible interventions on predicate $\mbox{\sf ERE}$ could be easily handled with a disjunctive rule like that in P3., but without the ``transition" annotation $\bstar$. However, the $\mbox{\sf ERE}$ record is commonly a component, or a sub-record, of a larger record containing all the feature values for an entity \cite{deem}. Hence, the need for a more general and uniform form of specification.

\red{Here we are considering the simple scenario in which the values are treated as unordered and categorical (binary) values. In some applications of bucketization and one-hot-encoding, one assumes and takes advantage of an underlying order inherited from the values in the buckets. Such an order could be adopted and brought into this framework by using additional rules that define that order. Developing the details is somehow outside the scope of this work.}

\section{\red{Beyond Binary Features and Uncertainty}}\label{sec:new}

\subsection{Expectation  over interventions for the \ {\sf x-Resp} score}\label{sec:exp}

The {\sf x-Resp} introduced in Section \ref{sec:causes} could be considered as a first approach to quantifying the relevance of a feature value. However, as the following example shows, we might want to go one step further.

\begin{example} Consider a simple entity  $E(\e;0,a_1)$, with $0 \in \nit{Dom}(F_1) = \{0,1\}$ and $a_1 \in \nit{Dom}(F_2) = \{a_1,  \ldots a_k\}$.  Assume that $\mc{C}(E(\e;0,a_1)) = 1 = \mc{C}(E(\e_i;0,a_i)) = \mc{C}(E(\e_j';1,a_j))$, for $1\leq i \leq k, \ 1\leq j \leq k-1$, but $\mc{C}(E(\e_k';1,a_k)) = 0$.

 We can see that changing only the original first feature value does not change the label provided by the classifier. Nor does additionally changing the second feature value, except when using the last possibly value, $a_k$, for $F_2$. In this case, $\mbox{\sf x-Resp}_{\e,F_1}(0) = \frac{1}{2}$, despite the fact that almost all interventions on the second feature value do not change the label.

 A similar phenomenon would appear if we had $\nit{Dom}(F_1) = \{b_1, ...,b_k\}$, with large $k$, and  $\mc{C}(E(\e;b_1,a_1)) = 1 = \mc{C}(E(\e_i;b_j,a_1))$, for $j=1, \ldots, k-1$, but $\mc{C}(E(\e_k;b_k,a_1)) =0$.
 In this case, the value $b_1$ is a counterfactual cause with explanation responsibility $1$, despite the fact that most of the interventions of $b_1$ do not switch the label.
\ A way to compensate for this could be taking the label average over all possible interventions.   \boxtheorem
\end{example}

\vspace{-4mm}In order to extend the definition of the $\mbox{\sf x-Resp}$ by considering all possible interventions, we may consider the average of counterfactual labels over a given population, which would assume all entities are equally likely.  In  more general terms, we may assume the underlying entity population, $\mc{E}$, has a probability distribution, $P$, which we can use to express the extended $\mbox{\sf x-Resp}$ in terms of expected values, as follows.

Consider $\e \in \mc{E}$, an  entity under classification, for which $L(\e) =1$, and a feature $F^\star \in \mc{F}$. Assume we have:
\begin{enumerate}
\item $\Gamma \ \subseteq \ \mc{F} \smallsetminus \{F^\star\}$, \ a set of features that may end up accompanying feature $F^\star$.
\item $\bar{w} = (w_F)_{F \in \Gamma}$,  \ $w_F \in \nit{Dom}(F)$, \  $w_F \neq \e_F$, i.e.  new values for features in $\Gamma$.
\item $\e' := \e[\Gamma := \bar{w}]$, i.e. reset $\e$'s values for $\Gamma$ as in $\bar{w}$.
\item $L(\e') = L(\e) = 1$,  i.e. there is no label change with $\bar{w}$ (but maybe with an extra change for $F^\star$, in next item).
\item There is $v \in \nit{Dom}(F^\star)$, with \ $v \neq F^\star(\e)$ and $\e'' := \e[\Gamma := \bar{w},F^\star:=v]$.
\end{enumerate}
As in Definition \ref{def:er} and the paragraph that follows it, \ if \ $L(\e)  \neq  L(\e'') = 0$, \ $F^\star(\e)$ is an {\em actual causal explanation} for $L(\e) =1$, with ``contingency set"  $\langle \Gamma, \e_\Gamma\rangle$, where $\e_\Gamma$ is the projection of $\e$ on $\Gamma$.

In order to define the ``local" responsibility score, make $v$ vary randomly under conditions 1.- 5.:
\newred{\begin{equation}
\mbox{\sf Resp}^{\!P}\!(\e,F^\star,\Gamma,\bar{w}) \ := \ \frac{L(\e') - \mathbb{E}[L(\e'')~|~\e''_{\mc{F}\smallsetminus \{F^\star\}} = \e'_{\mc{F}\smallsetminus \{F^\star\}}]}{1 + |\Gamma|}. \label{eq:local}
\end{equation}}
If, as so far,  label $1$ is what has to be explained, then $L(\e')=1$, and the numerator is a number between $0$ and $1$.  Here, $\Gamma$ is fixed.  Now we can minimize its size, obtaining the (generalized) responsibility score as the maximum local value; everything relative to  distribution $P$:
\newred{\begin{eqnarray}
\mbox{\sf Resp}^{\!P}_{\e,F^\star}\!(F^\star(\e)) \ &:=& \ \mbox{\large \nit{max}} \ \ \mbox{\sf Resp}^{\!P}\!(\e,F^\star,\Gamma,\bar{w}) \label{eq:global}\\
&&\mbox{\phantom{oo}}|\Gamma| \ \mbox{\nit{min.}} \ \mbox{(\ref{eq:local})} >0 \nonumber \\ &&\mbox{\phantom{oo}}{\langle \Gamma, \bar{w}\rangle \ \models \ \mbox{1.} - \mbox{4.}}\nonumber
\end{eqnarray}}
\red{This score was introduced, with less motivation and fewer details, and definitely not on a causal basis,  in \cite{deem},}  where experiments are shown, and different probability distributions are considered.

\subsection{Domain knowledge under uncertainty }\label{sec:semP}

Different probability distribution on the entity population $\mc{E}$ can be considered in the definition and computation of the generalized responsibility score (c.f. Section \ref{sec:exp}).
A natural choice is the {\em uniform distribution}, $P^u$, that gives equal probability, $\frac{1}{|\mc{E}|}$, to each entity in $\mc{E}$. Another natural distribution is the {\em product distribution}, $P^\times$, that is obtained, under the assumption of independence, as the product of given marginal distributions, $p_{_{F}}$, of the features $F \in \mc{F}$: \ $P^\times(f_1,\cdots, f_n) := \Pi_{_{F_i \in \mc{F}}} p_{_{F_i}}(f_i)$.

We can also assume we have a sample from the entity population $S \subseteq \mc{E}$ that is used as a proxy for the latter. In this case, the distributions above become {\em empirical distributions}. In the uniform case, it is given by: $\hat{P}^{u}(\e) := \frac{1}{|S|}$ if $\e \in S$, and $0$, otherwise. \ In the case of the product, $\hat{P}^\times(f_1,\ldots,f_n) := \Pi_{_{F_i \in \mc{F}}} \hat{p}_{_{F_i}}(f_i)$, with $\hat{p}_{_{F_i}}(f_i) := \frac{|\{\e \in S~|~ \e_i =f_i\}|}{|S|}$.  A discussion on the use of these distributions in the context of explanation scores can be found in \cite{deem}.

In general, one can say that the uniform distribution may not be appropriate for capturing   capturing correlations between feature values. One could argue that  certain combinations of feature values  may be more likely than others; or that certain correlations among them exist. This situation is aggravated by the product distribution due to the independence assumption.  For these reasons an empirical distribution may be better for this purpose.

In any way, having chosen a distribution on the population, $P^\star$, to work with; in particular, to compute the  expectations needed for the responsibility score in (\ref{eq:local}),  one could consider modifying  the probabilities in the hope of capturing correlations and logical relationships between feature values. In particular, one could introduce {\em constraints} that prohibit certain combinations of values, in the spirit of {\em denial constraints} in databases, but in this case admitting positive and negative atoms. For example,  with propositional features $\mbox{\nit{Old}}$ standing for {\em ``Is older than 20"} and $\mbox{\nit{OverDr}}$ for {\em ``Gets an account overdraft above \$50K"}, we may want to impose the prohibition $\neg (\overline{\mbox{\nit{Old}}} \wedge \nit{OverDr})$, standing for ``{\em nobody under 20 gets at overdraft above \$50K}".

    These constraints, which are satisfied or violated by a single entity at a time,  are of the form:
\begin{equation}
\chi: \ \neg (\bigwedge_{F \in \mc{F}_1} F \wedge \bigwedge_{F' \in \mc{F}_2} \bar{F'}), \label{eq:cons}
\end{equation}
where $\mc{F}_1 \cup \mc{F}_2 \subseteq \mc{F}$, \ $\mc{F}_1 \cap \mc{F}_2 = \emptyset$, and $F, \bar{F'}$ mean that features $F, F'$ take values $1$ and $0$, resp.

The event associated to  $\chi$ is \
$E(\chi) = \{\e \in \mc{E}~|~ \e \models \chi\}$,
where $\e \models \chi$ has the obvious meaning of satisfaction of $\chi$ by entity $\e$.
In order to accommodate the constraint,  given the initial  probability space $\langle \mc{E}, P^\star\rangle$, we can redefine the probability as follows. For $E \subseteq \mc{E}$,
\begin{equation}P^\star_\chi(E) := P^\star(E|E(\chi)) = \frac{P^\star(E \cap E(\chi))}{P^\star(E(\chi))}.\label{eq:newP}
\end{equation}
If $\chi$ is  consistent with the population, i.e. satisfiable in $\mc{E}$, the conditional distribution is well-defined.
 Now, the probability of $\chi$'s violation set is:
$$P^\star_\chi(\mc{E} \smallsetminus E(\chi)) = \frac{P^\star(\emptyset)}{P^\star(E(\chi))} = 0.$$
This definition can be extended to finite and consistent sets, $\Theta$, of constraints,  by using $P^\star_{\wedge\Theta}(E)$  in (\ref{eq:newP}), with $\wedge\Theta$ the conjunction of the constraints in $\Theta$.

Of course, one could go beyond constraints of the form (\ref{eq:cons}), applying the same ideas, and consider any propositional formula that is intended to be evaluated on a single entity at a time, as opposed to considering  combinations of feature values for   different entities.

The resulting modified distribution that accommodates the constraints could be used in the computation of any of the scores expressed in terms of expected values or in probabilistic terms.

An alternative approach consists in restricting the (combinations of) interventions in the definitions and computation of the responsibility score, as suggested in Section \ref{sec:sem} (and any other score based on counterfactual interventions, as a matter of fact). It is worth performing experimental comparisons between the two approaches.

\section{\red{Related Work}}\label{sec:relw}

In this worl we consider only {\em local methods} in that we are not trying to explain the overall behavior of a classifier, but a particular output on the basis of individual feature values. We also consider {\em model-agnostic methods} that can be applied to black-box models, i.e. without the need for an access to the internal components of the model. Of course, these approaches can be applied to open models, i.e. that make all its components transparent for analysis. Actually, in some cases, it is possible to take computational advantage of this additional source of knowledge  (more on this below in this section).

There are several approaches to the explanation of outcomes from classification models.  These methods can be roughly categorized as those that provide  {\em attribution scores} and those that provide {\em sufficient explanations}. Those in the former class assign a number to each feature value that reflects its relevance for the outcome at hand. $\mbox{\sf x-Resp}$ falls in this category. For comparison with other approaches, we repeat that, in the case
of $\mbox{\sf x-Resp}$, one counterfactually modifies feature values to see if the outcome changes. The score is computed from those changes.

Counterfactual changes are omnipresent in attribution-score-based methods, and they can be used to consider alternative entities to the one under explanation, and a notion of distance between those alternatives and the latter \cite{martens,wachter,russell,karimiA,datta,deem}. Sometimes the counterfactuals are less explicit, as with the popular $\mbox{\sf Shap}$ score \cite{lund20}, that is based on the Shapley value of game theory. It can be seen as a counterfactual-based score in that all possible combinations of features values (and then, most of the time departing from the initial entity) are considered in a complex aggregation (an average or expected value).

 The $\mbox{\sf Shap}$ score   is designed as a model-agnostic method. However, for a large class of classifiers whose internal components can be used for the score computation, $\mbox{\sf Shap}$ becomes computable in polynomial-time, while its general computational complexity is $\#P$-hard \cite{aaai21,guy}. As we showed in this paper, the computation of the $\mbox{\sf x-Resp}$ is $\nit{N\!P}$-hard. Actually, the generalized, probabilistic extension of $\mbox{\sf x-Resp}$ (c.f. Section \ref{sec:exp}), for certain classifiers and probability distributions on the underlying population, $\mbox{\sf x-Resp}$ can be $\#P$-hard \cite{deem}. An investigation of classes of classifiers for which the $\mbox{\sf x-Resp}$ score (deterministic as in this work or probabilistic) can be computed in polynomial time is still open.

The popular $\mbox{\sf LIME}$ score \cite{lime} is also an attribution score.  It appeals to an explainable model that locally approximates the agnostic model around the entity under classification. From the resulting approximation  feature scores can be computed.

Sufficient-explanation methods try to identify those (combinations of) feature values that alone determine the outcome at hand, in the sense that, by keeping those values and possibly changing all the others, the outcome remains the same \cite{anchors,guyNew}. One could say, in some sense, that the outcome is {\em entailed} by the values that appear in a sufficient explanation. Identifying those values is reminiscent of performing an abductive diagnosis task, as done with rule-based specifications \cite{eiter97}, actually \cite{anchors} does appeal to rule-based methods.

There are some approaches to logic-based explanations for classification. They mostly follow the {\em sufficient-explanation} paradigm we mentioned above. More specifically, it has been shown how to ``reconstruct" certain classes of classifiers, e.g. random forests, Bayesian classifiers, and binary neural networks, as Boolean circuits  \cite{shih18,shi20,Choi20}. Once a circuit is available, one can use it in particular to obtain explanations for the outcomes of the model using methods that are, in essence, abductive \cite{darwicheEcai20}. In this context the work presented in \cite{marquesAbd,ignatiev19,izza} is also relevant, in that logic-based encodings of neural networks, boosted trees, and random forests are proposed and exploited for explanation purposes. Abductive and SAT-based approaches are followed. Notice that abductive methods that generate sufficient explanations can also be the basis for score definitions and their computation. Just for the gist, if a feature value appears in the large number of sufficient explanations, then it could be assigned a large individual score.

To the best of our knowledge, none of the approaches described above, and others, are based on {\em logical specifications of the counterfactuals} involved in the score definition and computation. Furthermore, these are specifications that can easily adopt domain or desirable logical constraints in a seamless manner, and for combined use. Actually, the logic-based representations of complex classifiers that we just mentioned above, could be the starting point for the use of our approach. For example, a Boolean circuit can be represented as a set of rules that becomes a first component of a CIP that does the counterfactual analysis on that basis.

\ignore{
DeepLIFT    computes  importance  scores  for input  features  by  comparing  it  with  a  reference  entity.\\
Avanti Shrikumar, Peyton Greenside, and Anshul Kundaje.  Learning important features through propagating activation differences.  In ICML, pages 3145-3153, 2017.
}

\section{Discussion}\label{sec:disc}

This work is about interacting via ASP with possibly external classifiers, and reasoning about their potential inputs and outputs. The classifier is supposed to have been learned by some other means. In particular, this work is not about learning ASPs, which goes in quite a different direction \cite{russo}.

\red{In this work we have treated classifiers as black-boxes that are represented by external predicates in the ASP. However, we have also considered the case of a classifier that is specified within the CIP by  a set of rules,  to define the classification predicate $\mc{C}$. This was the case of a deterministic Decision Tree. Basically, each branch from the root to a label can be represented by a rule, with the branching direction at intermediate nodes represented by values in literals  in a rule body, and with the label in the rule head. \ Something similar can be done with Boolean Circuits used as classifiers. Actually, it is possible to represent more involved classifiers as Boolean circuits (c.f. Section \ref{sec:relw}).}

 \red{Our CIPs can be easily enhanced with different  extensions. For example, the feature domains can be automatically specified and computed from training or test data \cite{deem}, or other sources. As done in \cite{deem} for experimental purposes and using purely procedural approaches, it is possible in our ASP setting to restrict the sizes of the contingency sets, e.g. to be of size $2$ (c.f. Section \ref{sec:causes}). This can be easily done by adding a cardinality condition to the body of the disjunctive intervention rule (which is supported by {\em DLV-Complex} and {\em DLV2}). Doing this would not lead to any loss of best explanations (as long as they fall within the imposed bound),} and may reduce the computational work.

\red{ It is also possible to extend CIPs to make them specify and compute the {\em contingency sets} (of feature values) that accompany a particular value that has been counterfactually changed (c.f. Section \ref{sec:causes}). This requires a {\em set-building operation}, which is provided by {\em DLV-Complex}. \newred{Doing this would allow to obtain s-explanations, i.e. with minimal (not necessarily minimum) contingency sets (c.f. Example \ref{ex:nonMin}). One could try if, by deleting one element from the contingency set, the label changes or not. Again, {\em DLV-Complex} could be used here. In \cite{ijclr21} a detailed example can be found where this is illustrated at the light of the naive Bayes classifier.
This approach was also followed in \cite{foiks18} to compute contingency {\em sets} for individual database tuples as causes for query answers.} }

 Our specification of counterfactual explanations is in some sense {\em ideal}, in that the whole product space of the feature domains is considered, together with the applicability of the classifier over that space. This may be impractical or unrealistic. However, we see our proposal as a conceptual and generic specification basis that can be adapted in order to include more specific declarative practices and mechanisms.

 For example, restricting the product space can be done  in different manners. One can use constraints or additional conditions in rule bodies. A different approach consists in replacing the product space with the entities in a {\em data sample} $S \subseteq \Pi_{i=1}^n \nit{Dom}(F_i)$. We could even assume that this sample already comes with classification labels, i.e.
$S^L = \{\langle \e_1',L(\e_1')\rangle, \ldots, \langle \e_K',L(\e_K')\rangle\}$. This dataset may not be disjoint from the training dataset $T$ (c.f. Section \ref{sec:causes}). The definition of counterfactual explanation and CIPs could be adapted to these new setting without major difficulties.

The CIPs we have introduced are reminiscent of  {\em repair programs} that specify and compute the repairs of a database that fails to satisfy the intended integrity constraints \cite{monica}. Actually, the connection between database repairs and actual causality for query answers was established and exploited in \cite{tocs}. \red{ASPs that compute tuple-level and attribute-level causes for query answers were introduced in \cite{foiks18}. Attribute-level causes are close to interventions of feature values, but the ASPs for the former   are much simpler that those presented here, because in the database scenario,} changing attribute values by nulls is good enough to invalidate the query answer (the ``equivalent" in that scenario to switching the classification label here). Once a null is introduced, there is no need to take it into account anymore, and a single ``step" interventions are good enough.

\red{Our CIPs are designed to obtain general counterfactual explanations, and in particular and mainly, c-explanations. The latter are associated to minimum-size contingency sets of feature values, and, at the same time, to maximum-responsibility feature values. This is achieved via the weak constraints in (\ref{eq:weak}). If we wanted to obtain the responsibility of a non-maximum-responsibility feature value, that is associated to an s-explanation that is not a c-explanation,  we can remove the weak constraints (and by so doing keeping all the models of the CIP), and pose a query under the {\em brave semantics} about the values of $\mbox{\nit{inv-resp}}$ in (\ref{eq:m}). An approach like this was followed in \cite{foiks18} for database tuples as causes for query answers.}

\red{Apropos query answering, that we haven't exploited in this work, several opportunities offer themselves. For example, we could pose a query under the brave semantics to detect if a  particular feature value is ever counterfactually changed, or counterfactually changed in a best explanation. Under the skeptical semantics, we can identify feature values that change in all the counterfactual entities. Fully exploiting query answering is a matter of ongoing work.}

\red{In this work we have considered s- and c-explanations, that are associated to two specific and related minimization criteria.
 However, as done in abstract terms in Section \ref{sec:causes}, counterfactual explanations could be cast in terms of different optimization criteria \cite{karimiA,russell,max}. One could investigate  the specification and implementation of other forms of preference, the generic $\preceq$ in Definition \ref{def:causal:explanation}, by using ASPs as  in \cite{schaub,asp2}.}

 \red{Specifying and computing the generalized, probabilistic responsibility score of Section \ref{sec:exp} goes beyond the usual capabilities of ASP systems. However, it would be interesting to explore the use of {\em probabilistic} ASPs for these tasks \cite{pASP,lee}.  Along a similar line, {\em probabilistic} ASPs could be in principle used to deal with the integration of semantic constraints with underlying probability distributions on the entity population, as described in Section \ref{sec:semP}. This is all  matter of ongoing work.}

\vspace{2mm}
\noindent {\bf Acknowledgements:} \ The author has been a member of RelationalAI's Academic Network, which has been a source of inspiration for this work, and much more. Part of this work was funded by ANID - Millennium ScienceInitiative Program - Code ICN17002. Help from  Jessica Zangari and  Mario Alviano with information about  DLV2, and from Gabriela Reyes with the DLV program runs is much appreciated.

\end{document}